%% file: main.tex
\documentclass[10pt,twocolumn,letterpaper]{article}

\usepackage[pagenumbers]{cvpr} 

\usepackage{soul}
\usepackage{url}
\usepackage{graphicx}
\usepackage{wrapfig}
\usepackage{makecell}
\usepackage{wrapfig}
\usepackage{lipsum}
\usepackage{algpseudocode}
\usepackage{pdflscape} 
\usepackage{verbatim}
\usepackage{tabularx}

\usepackage{amsmath, amssymb, graphicx, xspace}
\usepackage{booktabs}
\usepackage{bbm} 
\usepackage{algorithm}

\newcommand{\xmark}{\texttimes}
\usepackage[pagebackref,breaklinks,colorlinks,allcolors=cvprblue]{hyperref}
\usepackage{pifont}
\definecolor{cvprblue}{rgb}{0.21,0.49,0.74}
\title{GFlowVLM: Enhancing Multi-step Reasoning in Vision-Language Models with Generative Flow Networks}

\author{Haoqiang Kang$^{1,}$$^{2*}$ \;\;
Enna Sachdeva$^1$ \;\;
Piyush Gupta$^1$ \;\;
Sangjae Bae$^{1}$\;\;
Kwonjoon Lee$^{1}$\\
{\normalfont
$^1$Honda Research Institute USA; $^2$University of California San Diego}
}
\def\paperID{5092} 
\def\confName{CVPR}
\def\confYear{2025}

\begin{document}

\maketitle

\def\thefootnote{$*$}\footnotetext{Work was done during Haoqiang Kang's internship at Honda Research Institute.}

\begin{abstract}
Vision-Language Models (VLMs) have recently shown promising advancements in sequential decision-making tasks through task-specific fine-tuning. However, common fine-tuning methods, such as Supervised Fine-Tuning (SFT) and Reinforcement Learning (RL) techniques like Proximal Policy Optimization (PPO), present notable limitations: SFT assumes Independent and Identically Distributed (IID) data, while PPO focuses on maximizing cumulative rewards. These limitations often restrict solution diversity and hinder generalization in multi-step reasoning tasks. To address these challenges, we introduce a novel framework, GFlowVLM, a framework that fine-tune VLMs using Generative Flow Networks (GFlowNets) to promote generation of diverse solutions for complex reasoning tasks. GFlowVLM models the environment as a non-Markovian decision process, allowing it to capture long-term dependencies essential for real-world applications. It takes observations and task descriptions as inputs to prompt chain-of-thought (CoT) reasoning which subsequently guides action selection. We use task based rewards to fine-tune VLM with GFlowNets. This approach enables VLMs to outperform prior fine-tuning methods, including SFT and RL. Empirical results demonstrate the effectiveness of GFlowVLM on complex tasks such as card games (NumberLine, BlackJack) and embodied planning tasks (ALFWorld), showing enhanced training efficiency, solution diversity, and stronger generalization capabilities across both in-distribution and out-of-distribution scenarios. Project page is available at \href{https://mk322.github.io/gflowvlm/}{https://mk322.github.io/gflowvlm/}.

\end{abstract}
\vspace{-12pt}
\section{Introduction}
Vision-Language Models (VLMs) have achieved remarkable results in generalized tasks such as image captioning and visual question answering \cite{blip,liu2024visual,minigpt4}. However, they struggle with structured reasoning in sequential decision making tasks that require causal understanding~\cite{chen2024vision}, especially in long horizon planning for tasks such as embodied AI, where agent must capture long term dependencies.

Recent advancements in LLMs and VLMs demonstrate emergent reasoning capabilities by leveraging Chain-of-Thought (CoT) reasoning, that enhances decision-making in multi-step interactive environments~\cite{wei2022chain, mu2024embodiedgpt, gupta2025generalized}. Typically, these models are fine-tuned using specialized visual instruction-following datasets through Supervised Fine Tuning (SFT) methods~\cite{liu2024visual, liu2024improved}, without active interaction with the environment, or optimized through Reinforcement Learning (RL) approaches, such as Proximal Policy Optimization (PPO)~\cite{schulman2017proximal, zhai2024fine}. However, SFT approaches often limits generalization to unseen scenarios, as training relies on maximizing the likelihood over a limited, specialized dataset, thereby restricting diversity in the solution space~\cite{li2023distilling}. Furthermore, RL methods like PPO tend to prioritize short-term rewards, which can hinder the model’s ability to consider long-term outcomes. Consequently, limited exploration in these models may lead to the oversight of more optimal long-term strategies, resulting in suboptimal performance in complex tasks~\cite{zhang2022proximal}.

In contrast to traditional reinforcement learning (RL) methods, which focus on maximizing cumulative rewards~\cite{gupta2022towards, gupta2022deterministic}, Generative Flow Networks (GFlowNets)~\cite{bengio2021flow} train stochastic policies to sample diverse, high-reward sequences (e.g., token sequences) with probabilities proportional to a specified reward function $R(x)$~\cite{bengio2023gflownet}. This approach samples sequences based on the reward function's distribution, enabling it to find a broader range of high-reward solutions beyond those typically identified by reward-maximizing techniques. Recent studies have applied GFlowNets to multi-step reasoning within the Large Language Models (LLMs) framework, demonstrating their effectiveness over maximum likelihood training and traditional reward-maximization methods~\cite{yu2024flow, takase2024gflownet, hu2023amortizing}. However, these methods lack the multimodal capabilities which are crucial for embodied AI tasks requiring the integration of visual and textual information. Additionally, related works, such as FoR~\cite{yu2024flow}, relies on Markovian structures within this framework, which may fail to capture the long-range dependencies necessary for complex, real-world reasoning tasks.

To address these limitations, we introduce \texttt{GFlowVLM}, a novel approach integrating GFlowNets with VLMs in an end-to-end fine-tuning framework. It explicitly models non-Markovian flows, enabling richer multimodal reasoning suitable for complex sequential decision-making. To our knowledge, \texttt{GFlowVLM} is the first to fuse GFlowNets with VLMs directly, addressing the distinct challenges posed by multimodal, sequential reasoning crucial in structured planning environments. Our approach initializes a policy with a pretrained VLM and fine-tunes it using GFlowNets, guiding VLMs toward structured reasoning processes that capture logical dependencies between successive states. By implicitly representing reasoning as a tree structure—where nodes correspond to states with prior actions and observations, and edges represent actions leading to the next state—\texttt{GFlowVLM} enhances efficient learning of diverse and complex reasoning sequences. Empirical results demonstrate that \texttt{GFlowVLM} outperforms standard fine-tuning techniques including SFT and RL methods including PPO, by enhancing structured multimodal reasoning capabilities.\\
Our main contributions are as follows:
\begin{itemize}

    \item  We introduce a novel framework that integrates GFlowNets with common-sense capabilities of VLMs for multi-step decision making tasks, enhancing their reasoning abilities. To the best of our knowledge, this is the first work to explore this integration.
    
    \item  By fine-tuning VLMs with GFlowNets, we improve their capacity to handle complex reasoning tasks, enabling better exploration of reasoning paths, generating diverse solutions, achieving stronger generalization to out of distribution tasks.
    
    \item Through extensive experimentation, we demonstrate that our framework achieves better training efficiency, higher success rate and diversity in solution generation tasks compared to existing methods.
    
\end{itemize}
\vspace{-6pt}
\section{Related Works}
\vspace{-1pt}
\paragraph{Multi-Step Reasoning with Vision Language Models} Recent research has advanced the reasoning capabilities of large foundation models through specialized prompting techniques \cite{wei2022chain, dong2022survey, yao2022react, yao2024tree, wang2023describe, xi2023rise, park2023generative} and fine-tuning methods \cite{szot2023large, mu2024embodiedgpt, chen2024vision} that often add MLP or transformer layers to frozen models to interface with action spaces. Reinforcement learning from human feedback (RLHF) also aids in developing reward models \cite{ouyang2022training}. The RL4VLM approach \cite{zhai2024fine} uses PPO to train VLMs but lacks the structured reasoning enabled by our GFlowNets method, which is designed for deeper understanding of complex tasks. VLM reasoning in interactive environments, particularly embodied AI, has gained attention \cite{mu2024embodiedgpt, zhai2024fine, wei2022chain, yao2023react}, but our GFlowNets approach uniquely enables structured reasoning, enhancing task comprehension.
\vspace{-5pt}
\paragraph{GFlowNets} GFlowNets \citep{bengio2021flow} were originally created to learn sampling policies from unnormalized distributions, primarily aiding scientific discovery by generating diverse, high-reward samples \citep{jain2023gflownets, shen2023tacogfn, jain2022biological}. They have since been applied in recommendation systems \citep{liu2023generative}, domain adaptation \citep{zhu2023generalized}, combinatorial optimization \citep{zhang2023let, kim2024ant}, and enhancing neural network interpretability \citep{li2023dag}. GFlowNets also support sampling from complex posteriors \citep{hu2023gflownet}, sparse reward RL \citep{pan2023better}, and multi-objective optimization \citep{jain2023multi}. Recent adaptations fine-tune LLMs for multi-step reasoning tasks \citep{yu2024flow}, yet lack the multimodal capability for embodied AI planning, which we address in this paper.

\vspace{-3pt}
\section{Preliminaries}
\paragraph{GFlowNets} GFlowNets are models that amortize the cost of sampling from a target distribution over terminal states $\mathcal{X}$ by learning an approximation of this distribution based on its reward function. Given a directed acyclic graph (DAG) $\mathcal{G} = (\mathcal{S}, \mathcal{A})$ with states $\mathcal{S}$ and directed actions $\mathcal{A}$, there is an initial state $s_0$ and terminal states $\mathcal{X} \subset \mathcal{S}$. A trajectory $\tau = (s_0 \to \ldots \to s_n = x)$ represents a complete sequence ending in a terminal state $x \in \mathcal{X}$. \label{preliminary}
The trajectory flow $F : \mathcal{T} \rightarrow \mathbb{R}_{\geq 0}$ defines flows over trajectories, with state flow $F(s) = \sum_{s \in \tau} F(\tau)$. A forward policy $P_F(\cdot|s)$, often parametrized by a neural network, induces a distribution over trajectories and a marginal distribution over terminal states, with probabilities given by: $P_F(\tau) = P_F(s_0 \to \ldots \to s_n) = \prod_{t=0}^{n-1} P_F(s_{t+1}|s_t) \quad \forall \tau \in \mathcal{T}$. Similarly, a backward policy $P_B(\tau) = P_B(s_n \to \ldots \to s_0) = \prod_{t=0}^{n-1} P_B(s_{t}|s_{t+1}) \quad \forall \tau \in \mathcal{T}$. Given a non-negative reward function $R : \mathcal{X} \rightarrow \mathbb{R}_{\geq 0}$, GFlowNets aim to estimate a policy where the likelihood of sampling $x \in \mathcal{X}$ is proportional to $R(x)$. Thus, there exists a constant $Z$ such that: $R(x) = Z \sum_{\tau=(s_0 \rightarrow \ldots \rightarrow s_n = x)} P_F(\tau)\quad  \forall x \in \mathcal{X}$, where $Z= F(s_0) = \sum_{\tau \in \mathcal{T}}  F(\tau)$ is total flow at the initial state. See~\cref{sec:gflownet_prel} for more details.
\vspace{-5pt}
\paragraph{Off-Policy Training} An advantage of GFlowNets is their ability to leverage off-policy training data by reusing transitions from past trajectories to update the forward policy $P_F$\cite{bengio2021flow}. Unlike on-policy reinforcement learning methods, GFlowNets handle diverse, multimodal distributions effectively, using off-policy samples to approximate $R(x)$. This approach improves sample efficiency and accelerates convergence, especially in settings where generating trajectories is costly or where leveraging prior data is beneficial.

\subsection{Motivating Experiment}

\begin{figure}
    \centering
    \includegraphics[width=0.7\linewidth]{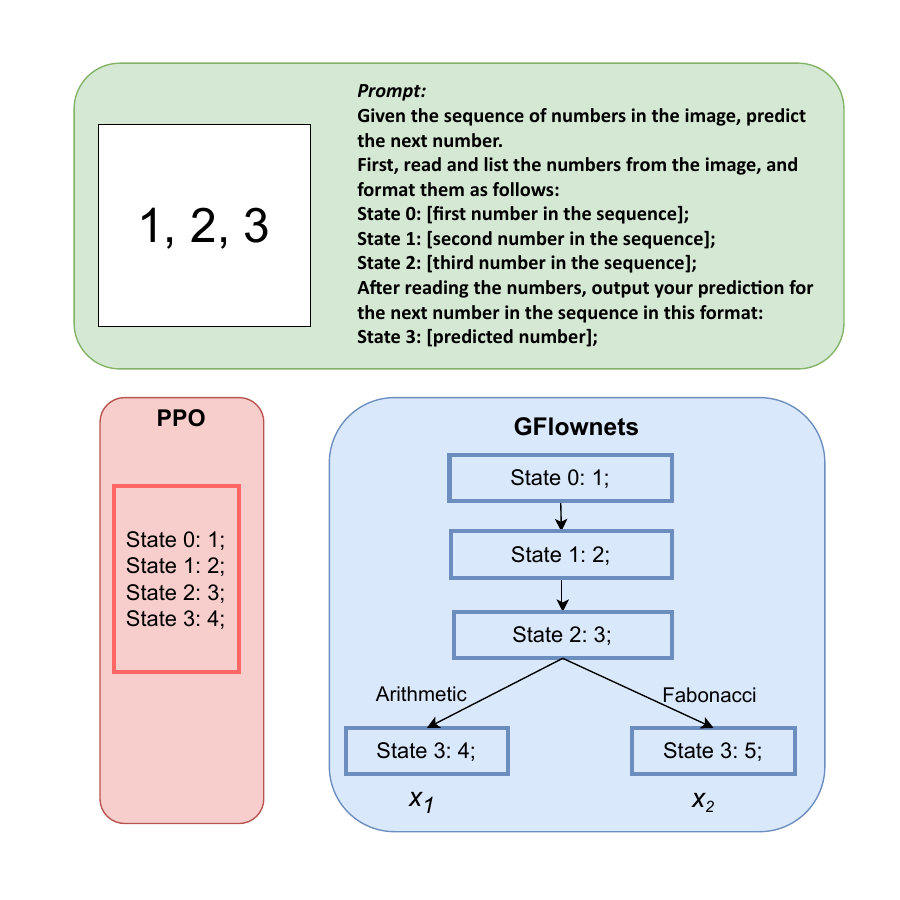}
    \caption{Overview of the prediction of diverse sequence using Gflownets as compared to PPO. The model takes the image of sequence and prompt as input, and generates the next number of sequence by implicitly modeling the causality. See~\cref{fig:teaser_alfworld} for a practical example.}
    \label{fig:teaser}
\end{figure}
\vspace{-3pt}

\begin{table}[ht!]
\small
\centering
\begin{tabular}{lccc}
\toprule
\textbf{Methods}  & \textbf{Temp. $\alpha$}  & \textbf{SR (\%)} & \textbf{\# Solutions} \\
\midrule
\hline
w/o fine tuning & 1  & 15.7 & 1.10 \\ 
w/o fine tuning & 1.2  & 16.1 & 1.12 \\ 
SFT & 1  & 21.7 & 1.03 \\ 
SFT & 1.2  & 22.0 & 1.09 \\ 
PPO & 1 & 50.2 & 1.13 \\ 
PPO & 1.2 & 49.8 & 1.15 \\ 
GFlowVLM & 1  & \textbf{76.4}  & \textbf{1.60} \\ 
GFlowVLM & 1.2  & \textbf{77.9}  &  \textbf{1.61}\\ 
\bottomrule
\end{tabular}
\caption{Results of motivating experiments. $\alpha$ denotes the temperature parameter of decoding. }
\label{tab:motivating}
\end{table}


To demonstrate the limitations of traditional approaches and highlight the dependencies captured by GFlowNets, we design a toy experiment combining two types of numerical sequences: $(i)$ the Fibonacci sequence, defined by $F(n) = F(n-1) + F(n-2)$, where each term is the sum of the previous two, and $(ii)$ an arithmetic sequence with a constant increment, $S(n) = S(n-1) + k$, where \( k \) is a fixed step size (e.g., \( k = 2 \) for sequences like \([2, 4, 6, \dots]\)). The task presents the model with an image of a partial sequence and a prompt (as shown in \cref{fig:teaser}), to predict the next number in the sequence. We evaluate the performance of fine-tuning VLM (LLAVA-v1.6-Mistral-7B~\citep{liu2024llava}) using SFT, PPO, and GFlowNets, with temperature parameters $\alpha = 1$ and $\alpha > 1$ to assess stochastic performance, as shown in ~\cref{tab:motivating}. Success rate (SR), measured as the percentage of correct next-number predictions across 1,000 samples, shows GFlowVLM outperform PPO by 26\% and generate 40\% more diverse solutions. Compared to SFT, GFlowVLM achieves a 54\% higher success rate and yield 59\% more diversity in responses, underscoring their strength in learning and generalizing causal structures. This advantage stems from GFlowNets' ability to infer underlying causal reasoning structure of sequence by sequentially sampling reasoning paths, in contrast to the limited diversity observed with SFT and PPO. While this toy example highlights key conceptual benefits of GFlowNets, we include a practical example in Sec.~\ref{app:motivating_ex} of Supplementary Material demonstrating how GFlowVLM can be applied to embodied AI tasks in ALFWorld, showcasing its real-world reasoning capabilities. This addition provides further evidence of the method’s utility beyond synthetic settings and illustrates its effectiveness in a more grounded, task-oriented scenario.

\vspace{-7pt}
\section{Methodology}
\begin{figure*}[hbt!] 
    \centering
    \includegraphics[width=0.8\linewidth]{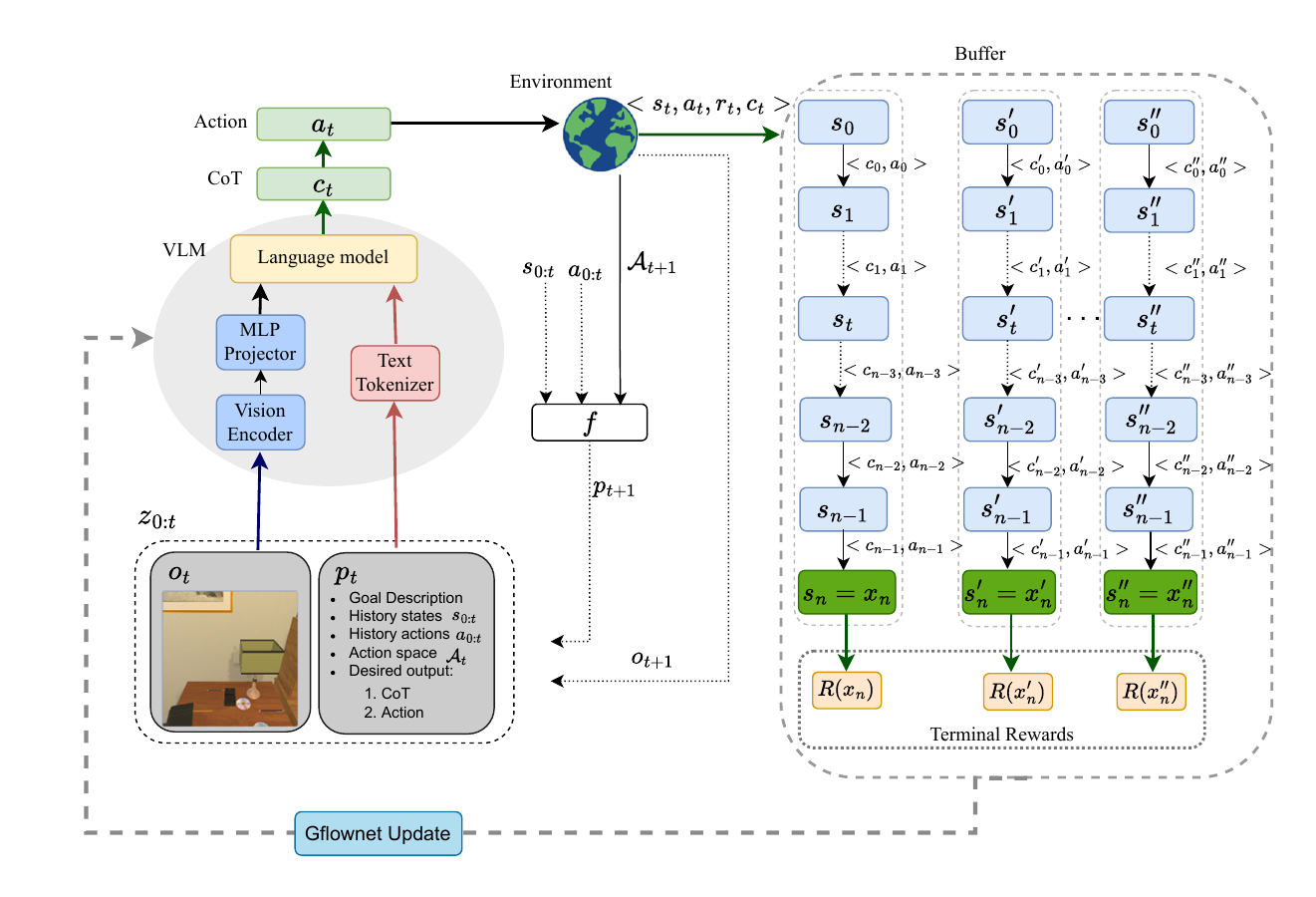}
       \caption{\textbf{Overall framework of proposed method:} The input $z_{0:t}$ at time step $t$ consists of a visual observation $o_t$ and an input prompt $p_t$ containing goal description, history states $s_{0:t}$, history actions $a_{0:t}$, and admissible actions $\mathcal{A}_{t}$, and outputs CoT reasoning $c_t$, and action $a_t$. The $a_t$ is executed in the environment to obtain reward $r_t(s_t, a_t)$, next observation $o_{t+1}$, and action space $\mathcal{A}_{t+1}$. $f$ generates the next prompt $p_{t+1}$ using description of next observation $o_{t+1}$ (if applicable), history of states $s_{0:t}$ and actions $a_{0:t}$ and next admissible actions $\mathcal{A}_{t+1}$. This generates multiple trajectories. The transitions $<s_t, a_t, r_t, c_t>$ , $<s_t', a_t', r_t', c_t'>$ and $<s_t'', a_t'', r_t'', c_t''>$ across different trajectories are added to buffer to update the forward policy $P_{F}$ using GFlowNets. $\{ x, x', x''\} \in \mathcal{X}$ represent the terminal states of sequences. $R(x)$ represents the non-negative reward obtained from the environment (after reward shaping, if applicable) at terminal state $x$ of a trajectory.}
    \label{fig:pipeline}
\end{figure*}
\vspace{-4pt}
This work utilizes GFlowNet’s structure learning to enhance the VLM's ability to obtain high-quality, diverse solutions whose distribution is proportional to the reward function. By fine tuning VLMs using GFlowNets, it allows the solutions to be sampled from the distribution of the reward function, which prevents learning policies settled around a small number of modes. \cref{fig:pipeline} shows the overall pipeline of our proposed framework. The model takes current observation image $o_t$ and designed task specific prompt $p_t$ as the input. $p_t$ contains the description of the goal, history actions $a_{1:t-1}$, history states $s_{1:t-1}$ and admissible action space corresponding to the current observation $o_t$. To incorporate non-Markovian assumption, input $z_{0:t}$ include history actions $a_{0:t}$ and states $s_{0:t}$, respectively along with the input image $o_t$\footnote{Only the current input image $o_t$ is used, as including every intermediate image would be computationally costly, and current VLMs do not perform optimally with multiple images as input.}. The desired output format includes the CoT reasoning $c_t$ and action $a_t$, where $a_t$ directly interacts with the environment.

\vspace{-5pt}
\subsection{VLM as a policy: Fine tuning VLMs using GFlowNets to estimate actions} 
\vspace{-2pt}
We use a non-Markovian approach, essential for reasoning tasks that depend on multiple past states to capture long-term dependencies, and tackle longer sequences—challenges that the Markovian assumption cannot adequately address. We fine tune VLM of LLaVA~\cite{liu2024llava} as a policy for structured reasoning, where VLM serves as the forward policy $P_F$, selecting the next action $a_t$ that advance the reasoning chain at every step $t$. For each task $\mathcal{W}$, the model takes the visual observation $o_t$ and prompt $p_t$ as inputs, and outputs the CoT and action. 
\vspace{-10pt}
\paragraph{Prompt Design} We retain the same prompt format as  ~\cite{zhai2024fine} for a fair comparison. However, to incorporate historical context in decision-making, we modify the prompt template to include the history of states and actions predicted by the VLM, as shown in~\cref{general_prompt_temp}. The textual prompt $p_t$ contains the goal description $g$, the history of states $s_{0:t}$ and actions $a_{0:t}$, and the action space $\mathcal{A}_{t+1}$ available after interacting with the environment. For certain tasks $q$ that may contain observation-dependent information, such as the textual description $d(o_{t+1})$ of the observation $o_{t+1}$, the function $f$ generates the prompt $p_{t+1}$ as: $p_{t+1} = f(d(o_{t+1})\cdot\mathbb{I}_{\{\text{Task} = \text{q}\}}, s_{0:t}, a_{0:t}, \mathcal{A}_{t+1})$, where $\mathbb{I}$ is an indicator function which is 1 only for a certain task $q$ if the observation-dependent information is available.
\vspace{-9pt}
\paragraph{Action Selection} Before selecting an action at each step $t$, we incorporate a CoT reasoning mechanism, where the model generates intermediate reasoning steps to guide the action selection process. At time $t$, the VLM first generates a reasoning CoT $c_t$, which includes a description of the image and intermediate thoughts. Since VLMs are pre-trained on large-scale image-caption data, CoT steps provide additional context and help the model explicitly consider dependencies between different states before selecting the next action. The CoT then guides the action selection. We define $z_t$ as the structured state at time $t$ in the trajectory, which includes the action $a_t$, environment state $s_t$, and visual observation $o_t$. More precisely, the trajectory $z_{0:t}$ is composed of the current visual observation $o_t$ and an input prompt $p_t$ that contains the goal description $g$, the history of environment states $s_{0:t-1}$, actions $a_{0:t-1}$, and the set of admissible actions $\mathcal{A}_t$. See~\cref{app:var-tb} for details.

The probabilities for the CoT and action sequences of tokens are defined as follows:
\begin{flalign}
P_{\text{CoT}}(c_t | z_{0:t},  g; \theta) =\\ \prod_{j=1}^{n_c} P_{\text{VLM}}(w_j | w_{<j}, z_{0:t},  g; \theta)
\end{flalign}
\vspace{-20pt}  
\begin{flalign}
P_{\text{Action}}(a_t | c_t, z_{0:t},  g; \theta) =\\ \prod_{i=1}^{n_a}  P_{\text{VLM}}(w_i | w_{<i}, c_t, z_{0:t},  g; \theta)
\end{flalign}
where \( n_c \) and \( n_a \) represent the number of tokens in the CoT sequence \( c_t \) and action sequence \( a_t \), respectively, and $w_i$ represents the $i$-th text token in a sequence. Here, \( P_{\text{VLM}}(w_i | w_{<i}, c_t, z_{0:t}, g; \theta) \) and \( P_{\text{VLM}}(w_j | w_{<j}, z_{0:t},  g; \theta) \) denote the VLM's token-level likelihoods for the action and CoT sequences, conditioned on previous tokens, the history of states \( z_{0:t} \), and goal description \( g \). The log forward policy $\log P_F(z_{t+1}| z_{0:t}, g; \theta)$ is then computed as a weighted sum of the log probabilities of CoT tokens $\log P_{\text{CoT}}(c_t | z_{0:t},  g; \theta)$, and the original log action probabilities $\log P_{\text{Action}}(a_t | z_{0:t},  g; \theta)$:

\vspace{-5pt}
\begin{flalign}
\begin{split}
\log P_F(z_{t+1}| z_{0:t}, g; \theta) = \log P_{\text{Action}}(a_t | z_{0:t},  c_t, g; \theta) +  \\
\lambda \log P_{\text{CoT}}(c_t | z_{0:t},  g; \theta),
\end{split}
\label{cot-factor-equ}
\end{flalign}
where $\lambda \in [0, 1]$ is a weighting factor that controls the influence of the CoT reasoning on the final action selection. The CoT probabilities $P_{\text{CoT}}(c_t | z_{0:t},  g; \theta)$ provide a structured, intermediate reasoning context that refines the decision-making process, ensuring that the final action is selected with consideration of both direct state information and the model's internal thought process. We perform an ablation study on the effect of $\lambda$, and we selected $\lambda = 0.4$ in our work, as discussed in the Sec. \ref{section-cot-ablation} of Supplementary Material.


\begin{table}[hbtp]
    \small
    \begin{tabularx}{0.48\textwidth}{X}
    \toprule
    \textbf{CoT prompt $p_t$ for task $\mathcal{M}$} \\ 
    You are trying to solve a task $\mathcal{M}$. \{Description of the task\}. The action space of $\mathcal{M}$ is \{all legal actions $a \in \mathcal{A}$\}. Use \texttt{[DONE]} when you think you have completed the task. \\
    Task: \{Task description\} \\
    State 0: \{Initial observation\} \\
    Action 0: \{First action\} \\
    State 1: \{Observation for step 1\} \\
    Admissible Next Actions: \{``action1'', ``action2'', `` \texttt{[DONE]}'' (if applicable)\} \\
    \\
    Your response should be a valid JSON file in the following format: \\
    \{
    \textcolor{PineGreen}{``thoughts'': ``{first describe what you see in the image using the text description, then carefully think about which action to take to complete the task.}''},     \\
    \textcolor{purple}{``action'': ``{an admissible action.}''}
    \}\\
    \hline
    \textbf{Formatted text output} \\
    \{ 
    \textcolor{PineGreen}{``thoughts'': ``Given the current state and previous steps, I should choose [$a_t$] as the next action.''}, \\
    \textcolor{purple}{``action'': ``$a_t$''} 
    \}\\
    \bottomrule
    
    \end{tabularx}
    
    \caption{A template showing the input prompt and corresponding output. The \textcolor{PineGreen}{green} text highlights the chain-of-thought reasoning which may contain task-specific descriptions, while the \textcolor{purple}{red} text indicates the action based on the description. }
    \label{general_prompt_temp}
\end{table}

\vspace{-8pt}
\subsection{Training Objectives}
We adopt three different objective functions of GFlowNets, \textit{Variance Trajectory-Balanced (TB)}~\cite{malkin2022trajectory}, \textit{Subtrajectory-Balanced (SubTB)}~\cite{madan2023learning}, and \textit{Detailed-Balanced (DB)}~\cite{bengio2023gflownet}, to finetune a VLM. We define $z_t$ as the state in the trajectory sequence that includes both $a_{t}$, $s_t$, and $o_t$. See the Sec. \ref{app:loss_fn} of the Supplementary Material for more details.


\vspace{-2pt}

\subsubsection{Variance Trajectory Balanced (Var-TB) Loss}

The \textit{Trajectory-Balanced} (TB) objective ensures that the probability of generating a complete trajectory $\tau = (z_0 \rightarrow z_1 \rightarrow \cdots \rightarrow z_n = x)$ is proportional to the reward $R(x)$. This objective is given by: 
\vspace{-8pt}







\begin{equation}
\mathcal{L}_{\mathrm{VarTB}}(\boldsymbol{\tau}; \theta) = \frac{1}{K} \sum_{k=1}^{K} \bigg(\mathrm{\zeta}(\tau_k; \theta)- \mathbb{E}_{\boldsymbol{\tau}}\big[\mathrm{\zeta}(\boldsymbol{\tau}; \theta)\big] \bigg)^2 ,
\end{equation}



where $\zeta(\cdot)$ is the estimated initial flow (see ~\cref{est. init flow} in Supplementary Material for details), $\tau_k$ is $k^{th}$ sampled trajectory during training, and $K$ represents the total number of trajectories. This loss ensures that the high-reward trajectories are sampled more frequently by the policy. 

\vspace{-3pt}
\subsubsection{Subtrajectory Balanced (SubTB) Loss}
The \textit{Subtrajectory-Balanced} (SubTB) loss operates on subtrajectories of the form $z_{0:m} = (z_0 \rightarrow z_1 \rightarrow \cdots \rightarrow z_m)$. It ensures that each segment of the reasoning path or structure remains consistent, where the flows are balanced locally between forward and backward transitions. We use a modified version of SubTB loss \cite{hu2023amortizing} as follows:
\vspace{-5pt}



\begin{flalign}
\begin{split}
\mathcal{L}_{\mathrm{SubTB}}(z_{0:m},g ; \theta) =  \sum_{0 \leq i < j \leq m} \\ \Bigg(\log \frac{R(z_{0:i}\top) \prod_{k=i+1}^{j} P_F(z_k | z_{0:k-1} ,g ; \theta) P_F(\top | z_{0:j} ,g ; \theta)}  {R(z_{0:j}\top)  P_F(\top | z_{0:i} ,g ; \theta)}\bigg)^2
\end{split}
\end{flalign}

where $i$ and $j$ are two time steps along a subtrajectory, and $\top$ is the \texttt{[DONE]} symbol to terminate the trajectory. $\top$ is generated similar to \cite{hu2023amortizing}. The reward $R(z_{0:i}\top)$ is computed as a cumulative reward given by the environment from step 0 to step $i$. This loss penalizes discrepancies in local transitions and ensures that all subsegments of a trajectory follow the correct balance conditions, reducing variance in smaller parts of the trajectory.

\vspace{-3pt}

\subsubsection{Detailed Balanced (DB) Loss}
The \textit{Detailed-Balanced} (DB) loss is used to ensure that each transition $z_t \rightarrow z_{t+1}$ between two states is balanced by matching the forward and backward flows at every step of the trajectory. Since it takes transition as input, we need dense rewards. The DB loss is formulated as:
\vspace{-5pt}


\begin{flalign}
\begin{split}
\mathcal{L}_{\mathrm{DB}}(z_{0:t} \rightarrow z_{t+1},g ; \theta) = \\ \Bigg(\log \frac{R(z_{0:t}\top) P_F(z_{t+1} | z_{0:t} ,g ; \theta) P_F(\top | z_{0:t+1} ,g ; \theta)}  {R(z_{0:t+1}\top)  P_F(\top | z_{0:t} ,g ; \theta)}\bigg)^2.
\end{split}
\end{flalign}
This loss ensures that every state-to-state transition follows the correct flow, preventing inconsistencies in the trajectory construction.
\vspace{-15pt}  
\paragraph{Remark} One challenge when implementing both the SubTB and DB losses is accurately estimating the termination probability, $P_F(\top | z_{0:t}, g ; \theta)$, which represents the likelihood of reaching a terminal state at any point in the trajectory. Incorrect estimation of this probability can lead to suboptimal training and unbalanced flows. To address this, we introduce a new token, \texttt{[DONE]}, into the tokenizer to explicitly model the terminal state, and use distinct prompt designs as shown in the ~\cref{general_prompt_temp}. Moreover, we perform an additional SFT step on correctly labeled examples before applying GFlowNets training. This initialization helps the model better estimate termination probabilities, resulting in improved overall performance (See ablation study in~\cref{section-sft-ablation}).

\begin{algorithm}[t]
\caption{Training VLM with GFlowNets}
\label{alg:training_vlm_gflownets}
\begin{algorithmic}
\small
\State \textbf{Input:} An environment \texttt{env}, an initial VLM with parameters $\theta_0$, a CoT reasoning scaling factor as $\lambda$, maximum episode length as $T$, number of tasks as $W$, number of collected trajectories per task as $K$.
\For{$w = 1, \dots, W$} 
     \State $\mathcal{B}_w = \emptyset$ 
    \For{$k = 1, \dots, K$} 
        \State $t = 0$ 
        \State $g, o_t, \mathcal{A}_{t} = \texttt{env.reset()}$ 
        \State $p_t= f(o_{t}, \mathcal{A}_{t})$

        \While{$t \leq T$}
            \State  $z_{0:t} = \langle o_{t}, p_{t} \rangle$

            \State $c_t, a_t = argmax P_{F}(z_{t+1}|z_{0:t}, g;\theta_{w-1})$ 

            \State $r_t, o_{t+1}, \mathcal{A}_{t+1} = \texttt{env.step}(a_t)$ 

            \State $\mathcal{B}_w = \mathcal{B}_w \cup \{(s_t, c_t, a_t, r_t\}$ 
          \State $p_{t+1} = f\big(d(o_{t+1}) \cdot \mathbb{I}_{\{\text{q}\}},$
        \State \hspace{2em} $s_{0:t}, a_{0:t}, \mathcal{A}_{t+1}\big)$


            \State $t = t + 1$
            \If{$t = T$ \textbf{or} task $w$ \text{is completed}}
                \State \textbf{break} 
            \EndIf
        \EndWhile
    \EndFor
    \State Update $\theta_{w-1}$ on the collected trajectories $\mathcal{B}_w$ for task $w$ to obtain $\theta_w$

\EndFor
\State \textbf{Output:} Updated parameters $\theta_W$ after $W$ tasks.
\end{algorithmic}
\end{algorithm}

\vspace{-5pt}
\vspace{-3pt}
\section{Experiments}
\vspace{-3pt}

We evaluate the performance of GFlowVLM on three distinct tasks that require multi-step visual-language reasoning. The Numberline and Blackjack tasks assess GFlowVLM's arithmetic reasoning capabilities with maximum steps set as 10, while Alfworld focuses on decision-making tasks that demand visual-semantic understanding with max steps set as 35 in a sequence. We mainly compare the performance with RL4VLM \cite{zhai2024fine} and SFT methods. For fair comparison, we use the same base VLM of LLAVA-v1.6-Mistral-7B~\citep{liu2024llava}. We conduct 4 independent runs with different random seeds, reporting mean and standard deviation. Episode success rate measures reasoning performance across tasks, while the diversity metric (\text{Div@N})~\cite{yu2024flow} quantifies unique correct solutions across N samples. The minimum for \textit{Div@N} is 1. NumberLine and Blackjack have discrete negative rewards, but GFlowNets inherently do \textit{not} support negative rewards (Sec.
\ref{preliminary}). Thus, we apply reward shaping as outlined in \cref{sec:environment} on these two tasks. 


\vspace{-3pt}
\subsection{Baselines}
\paragraph{SFT} We employ two versions of SFT in our baseline: SFT-w/o-\texttt{[DONE]} and SFT-w/-\texttt{[DONE]}. SFT-w/o-\texttt{[DONE]} uses the same GPT-4o dataset as in~\cite{zhai2024fine}. For SFT-w/-\texttt{[DONE]}, we include the \texttt{[DONE]} action in the training inputs and add correct examples where outputs explicitly contain \texttt{[DONE]}. We fine-tune the \textit{LLaVA-1.6-7B} model on this dataset for 1 epoch using the official script. To ensure consistency, downstream GFlowNets training for both SubTB and DB losses starts from the same SFT checkpoint that includes \texttt{[DONE]}.

\vspace{-12pt}
\paragraph{RL4VLM} We compare with RL4VLM~\cite{zhai2024fine}, which uses PPO to fine-tune the VLM. RL4VLM follows the same environment reward scheme as used in the GymCards tasks, where rewards are set to \([0, -1, 1]\). Additionally, it employs a Markovian approach by excluding history information from the prompt. To ensure a fair comparison, we modify the original setup with two additional configurations: one that replaces the default environment reward function with our custom reward function, and another that includes history information in the prompt in a non-Markovian manner. These adjustments allow us to evaluate the model’s performance under different reward functions and prompt history settings.



\begin{table*}[h!]
\small
\centering
\begin{tabular}{lcccccc}
\toprule
\textbf{Method} & \textbf{Train Data} & \textbf{Assump.} & \textbf{SFT Init.} & \textbf{NL} & \textbf{NL-OOD} & \textbf{BJ} \\ 
\midrule
\hline
SFT-w/o-\texttt{[DONE]}& Off & - & - & 24.8 & 0.0 & 23.1 \\ 
SFT-w/-\texttt{[DONE]} & Off & - & - & 24.0 & 0.0 & 20.2 \\ 
RL4VLM~\cite{zhai2024fine} & On & M  & \checkmark & 89.4 & 3.1 & 40.2 \\
RL4VLM~\cite{zhai2024fine}$^\dagger$ & On & NM  & \checkmark & 90.3 & 4.4 & 41.0 \\ 
RL4VLM~\cite{zhai2024fine}$^*$ & On & M & \checkmark & 34.8 & 1.9 & 23.5 \\ 
GFlowVLM w/ Var-TB  & On & NM & \checkmark & \textbf{100.0} & 6.2 & 41.4 \\ 
GFlowVLM w/ SubTB & On & NM & \checkmark& \textbf{100.0} & 7.0 & 41.7 \\ 
GFlowVLM w/ DB & On & NM & \checkmark& \textbf{100.0} & 9.1 & 42.2 \\ 
\midrule
\hline
\multicolumn{6}{c}{\textbf{Ablations - w/ Off-Policy Training data}} \\
GFlowVLM w/ Var-TB & Off & NM & \checkmark & \textbf{100.0} & 17.3 & 43.0 \\ 
GFlowVLM w/ SubTB & Off & NM & \checkmark & \textbf{100.0} & 16.7 & 42.4 \\ 
GFlowVLM w/ DB & Off & NM & \checkmark & \textbf{100.0} & \textbf{18.6} & \textbf{43.8} \\ 
\midrule
\hline
\multicolumn{6}{c}{\textbf{Ablations - w/o SFT Initialization}} \\
GFlowVLM w/ SubTB & On & NM & \xmark & 23.0 & 0.0 & 8.4 \\ 
GFlowVLM w/ DB & On & NM & \xmark & 24.3 & 0.0& 6.8 \\ 
GFlowVLM w/ SubTB & Off & NM & \xmark& \textbf{34.4} & 0.0 & \textbf{17.4} \\ 
GFlowVLM w/ DB & Off & NM & \xmark& 33.1 & 0.0& 13.8 \\ 
\bottomrule
\end{tabular}
\caption{Performance comparisons across baseline models for NumberLine (NL) and BlackJack (BJ) tasks  for in-distribution and out-of-distributions (OOD) tasks. $^*$We use the same reward function as ours. $^\dagger$We use the same prompt as ours to include history information for non-Markovian setting. NL-OOD stands for Number line with out-of-distribution tasks. On and Off represent On-Policy and Off-Policy, respectively. M and NM stands for Markovian and non-Markovian assumption respectively.}
\label{gym-results}
\end{table*}

\vspace{-3pt}
\subsection{Environments}
\label{sec:environment}
\textbf{Numberline}. This task involves moving a current number ``current: $y_t$'' to a ``target: $c$''. The agent’s goal is to align $y_t$ with $c$ by outputting an action $a_t$ from the discrete set \{``+'', ``-'', \texttt{[DONE]}(if applicable)\}. In-distribution examples include numbers from 0 to 5, and OOD examples range from 10 to 50. We revise the reward function to replace the original discrete rewards of -1, 0, and 1 with non-negative values as follows: $R(x) = R(c, y_t) = \frac{l}{|c - y_t| + 1}$, where $l$ is a scaling constant set to 100. This reward incentivizes the model to bring the current number closer to the target, progressively increasing the reward as the gap decreases. For fair comparison, we run RL4VLM~\cite{zhai2024fine} with revised reward structure. \\
\vspace{-2pt}\textbf{Blackjack}.\label{sec:blackjack} The Blackjack task requires VLM to reason with visual information and adapt to stochastic outcomes. The agent aims to win by selecting an action $a_t$ from \{``stand'', ``hit'', \texttt{[DONE]}(if applicable)\}. We revise the reward function to replace the original discrete rewards of -1, 0, and 1 with non-negative values as follows: $R(x) = \max(1 \times 10^{-10}, (r(x) + 1)\times 10)$, where $r(x)$ represents the environment's original reward for terminal state $x$. This scales the rewards and ensures they are strictly non-negative. For fair comparison, we run RL4VLM~\cite{zhai2024fine} with revised reward structure. \\
\textbf{ALFWorld}.\label{sec:alfworld} ALFWorld~\cite{shridhar2020alfworld} is an embodied AI environment combining a text-based interactive setup with a vision-language planning dataset. ALFWorld has a state-dependent action space $\mathcal{A}_t$; Our prompt instructs the VLM to choose from the admissible actions $\mathcal{A}_t$, and we evaluate out-of-distribution (OOD) performance using a test set of previously unseen scenes. We use the same non-negative reward function as used in \cite{zhai2024fine}, making it suitable for Var-TB and SubTB losses. However, since it lacks dense rewards for every transition, DB does not perform effectively.

\begin{table*}[h]
\small
\centering
\begin{tabular}{lcccccccccc}
\toprule
\textbf{Method} & \textbf{Assump.} & \textbf{Pick} & \textbf{Look} & \textbf{Clean} & \textbf{Heat} & \textbf{Cool} & \textbf{Pick2} & \textbf{Avg. } & \textbf{OOD} & \textbf{Div@16} \\ 
\midrule
\hline
SFT-w/o-\texttt{[DONE]}   & - & 39.2 & 0    & \textbf{14.4} & 11.1  & 0     & \textbf{28.6} & 17.1 & 3.3  & 1.06 \\ 
SFT-w/-\texttt{[DONE]}  & - & 32.7 & 0    & 10.3 & 10.8  & 0     & 21.8 & 15.9 & 3.0  & 1.02 \\ 
RL4VLM~\cite{zhai2024fine} & M  & 47.4 & 14.7 & 10.4 & 14.4  & 18.8  & 18.0 & 21.7 & 4.8  & 1.12 \\ 
RL4VLM~\cite{zhai2024fine} & NM & 49.1 & 13.5 & 9.8 & 15.2 & 20.1 & 20.6 & 22.1 &  6.1 & 1.11 \\ 
GFlowVLM w/ SubTB  & NM  & \textbf{50.0} & \textbf{23.1} & 10.0 & \textbf{18.7}  & \textbf{24.3}  & 23.7 & \textbf{26.1} & \textbf{12.3} & 1.40 \\ 
GFlowVLM w/ Var-TB  & NM  & \textbf{50.0} & 22.2 & 10.2 & 16.1  & 22.7  & 21.9 & 25.7 & 10.9 & \textbf{1.41} \\ 
\bottomrule
\end{tabular}
\caption{Results of ALFWorld. Since Alfworld does not provide dense rewards, we can not not using DB loss here. Furthermore, while RL4VLM and GFVLM with SubTB are trained with SFT initialization, GFVLM with TB-Var is without STF initialization since we do not need to model the flow.}
\label{tab:alf-results}
\end{table*}

\vspace{-5pt}

\section{Results Analysis}
\vspace{-2pt}
\textbf{Improved VLM Reasoning abilities on In-distribution samples}. Our experiments show that GFlowVLM significantly enhances VLM reasoning in tasks like NumberLine, Blackjack, and ALFWorld. As shown in ~\cref{gym-results}, it improves success rates on in-distribution examples by 12\% over RL4VLM, with an 8\% gain in Blackjack due to high-quality, off-policy trajectories. For ALFWorld, GFlowVLM achieves a 29\% success rate improvement (~\cref{tab:alf-results}), highlighting GFlowNets' role in generating accurate trajectories crucial for VLM reasoning.
\vspace{-10pt}
\paragraph{Diverse Solutions}
Our method generates more diverse solutions than other baselines. In ALFWorld, GFlowNets achieve 25\% and 33\% higher diversity than RL4VLM and SFT (\cref{tab:alf-results}), as measured by diversity metric, capturing a wider range of plausible solutions, offering a distinct advantage in scenarios that benefit from broader strategy exploration. In contrast to PPO, which optimizes a single best policy for long-term planning, GFlowNet finds multiple diverse high-reward solutions, making it better suited for structured generation (see Section~\ref{app:qual-results} for detailed qualitative results).

\vspace{-10pt}
\paragraph{Improved Generalization on OOD samples} 
Our method enhances VLM reasoning on OOD examples, with GFlowNets achieving higher OOD success rates than RL4VLM in NumberLine and ALFWorld tasks by 322\% and 156\%, respectively. This demonstrates GFlowNets' capacity for robust generalization through diverse, accurate trajectory sampling, enabling effective handling of complex, unseen scenarios.

\begin{figure*}[htbp!]
    \small
    \centering
    \includegraphics[width=0.9\linewidth, height=0.2\textheight]{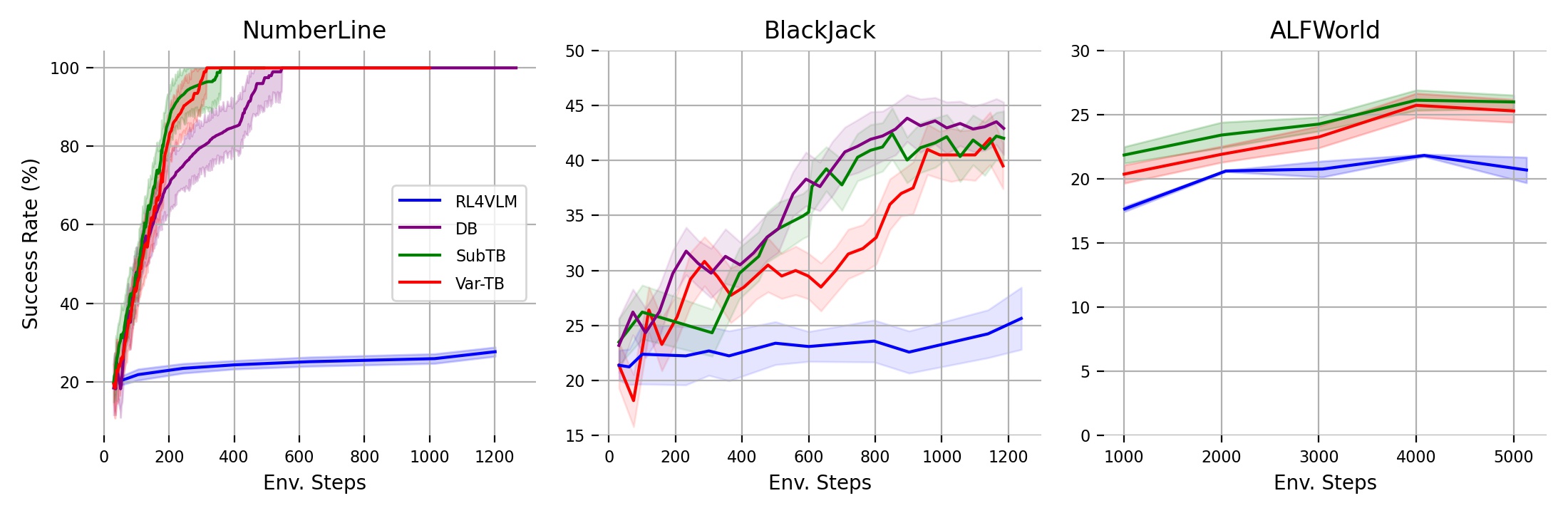}
    \caption{Training curves showing in-distribution episode success rates ($\%$) across three tasks. For Numberline and BlackJack, RL4VLM is trained with the original reward, while GFlowVLM variants use a revised reward function, as RL4VLM serves as a strong baseline under original rewards. In ALFWorld, all methods use the same (original) reward without revision. Models are trained using on-policy sampling.}
    \label{fig:training_curve}
\end{figure*}
\vspace{-10pt}
\paragraph{Benefits from Off-policy data} Since GFlowNets allow for off-policy~\cite{yu2024flow, malkin2022gflownets, kim2024improved} along with on-policy learning unlike PPO \cite{schulman2017proximal}, we adopt an off-policy data generation approach to evaluate the impact of using more accurate trajectories during training (see Section \ref{app_off_policy} for the details of data generation). \cref{gym-results} shows results for NumberLine (NL), out-of-distribution NumberLine (NL-OOD), and BlackJack (BJ) tasks. GFlowVLM with Var-TB, SubTB, and DB, demonstrate improvements with offline data, averaging a 36.2\% performance increase over online-only approaches. These results indicate that each loss function benefits from off-policy high-quality data, leveraging both correct and incorrect solutions to enhance performance, even in challenging OOD scenarios. \\
\vspace{-2pt}\textbf{Training Efficiency}. As shown in ~\cref{fig:training_curve}, GFlowNets converge faster than RL4VLM on NumberLine, Blackjack, and ALFWorld, reaching optimal performance with significantly fewer environment steps—about 10,000 fewer than RL4VLM. This efficiency reduces training time and computational demands, supporting scalability for complex reasoning tasks. \\
\vspace{-2pt}\textbf{Comparisons with different loss functions}. As shown in the training curves, all three loss functions—DB, Var-TB, and SubTB-converge at a similar rate. DB, which requires dense rewards for every transition since it utilize  transitions as input, demonstrates the best generalization, as evidenced in ~\cref{gym-results}, in the NumberLine and Blackjack tasks. Both SubTB and TB achieve comparable performance in terms of in-distribution and OOD generalization, making them equally effective for a wide range of reasoning tasks.\\
\textbf{SFT Initialization for SubTB and DB Losses}.\label{section-sft-ablation} The termination probability \(P_F(\top|\cdot)\) in DB and SubTB losses estimates the modified flow in GFlowNet~\cite{pan2023better}, which Var-TB lacks. To enable VLMs to accurately model this for DB and SubTB, we first apply SFT on correctly completed trajectories before fine tuning with GFlowNets. As shown in \cref{gym-results}, SFT initialization significantly boosts SubTB and DB performance on the NumberLine and Blackjack tasks. Without SFT, both losses perform poorly, especially on NL and BJ tasks. With SFT, SubTB and DB improve by 50\% and 36\% for NumberLine and by 107\% and 103\% for Blackjack, largely due to better estimation of terminal probability $P_F(\top |\cdot)$. \\
\textbf{Markovian and non-Markovian assumptions}. GFlowVLM outperforms RL4VLM in non-Markovian settings, excelling in complex, long-horizon tasks. In ALFWorld (\cref{tab:alf-results}), GFlowVLM achieves higher average performance by 18\%, OOD robustness by 100\%, and diversity by 27\%. It also achieves better success rates in gym tasks (\cref{gym-results}), where history aids decision-making, underscoring GFlowNets' advantage over PPO-based methods.





\vspace{-7pt}
\section{Conclusion, Limitation, Future Works}
\vspace{-3pt}
We introduce a novel framework using GFlowNets to enhance structured reasoning in VLMs, to capture relationships among reasoning steps for improved generalization. Unlike traditional methods like SFT and PPO, which are limited by certain assumptions, our approach supports complex, long-term reasoning tasks. Experiments in card games and embodied planning showed enhanced training efficiency, diversity, and generalization. We focus on a single-agent task setting, leaving multi-agent task and alternative prompting methods as future directions. Limited computational resources led us to use small-sized VLMs, but larger models may further benefit from GFlowNets.

\bibliographystyle{ieeenat_fullname}
\bibliography{main}

\clearpage

\appendix

\input{supp.tex}



\end{document}

%% file: supp.tex
\section{Preliminaries}
\label{sec:gflownet_prel}

\subsection{GFlowNets} 
We summarize the necessary preliminaries of GflowNets and encourage readers to refer to~\cite{bengio2023gflownet} for deeper understanding. In a directed acyclic graph $\mathcal{G} = (\mathcal{S}, \mathcal{A})$ with states $\mathcal{S}$ and directed actions $\mathcal{A}$, a complete trajectory is any trajectory starting in initial state $s_0$ and ending in terminal state $x \in X$ where $ X \subset \mathcal{S}$. There is a unique initial state $s_0 \in S$ with no parents. States with no children are called terminal, and the set of terminal states is denoted by $\mathcal{X}$. A trajectory $\tau = (s_0 \to \ldots \to s_n = x)$ represents a complete sequence ending in a terminal state $x \in \mathcal{X}$ where each $(s_t \to s_{t+1})$ is an action. The trajectory flow $F : \mathcal{T} \rightarrow \mathbb{R}_{\geq 0}$ defines flows over trajectories, with state flow $F(s) = \sum_{s \in \tau} F(\tau)$ and with edge flow $F(s \to s') = \sum_{\tau =(\ldots \to s \to s' \to \ldots)} F(\tau)$. The trajectory flow $F$ is Markovian if there exist action distributions $P_F(\cdot|s)$ over the children of each non-terminal state $s$.


\subsubsection{Forward and Backward Policies}
A forward policy $P_F(\cdot|s)$, often parametrized by a neural network, induces a distribution over trajectories and a marginal distribution over the children of every non-terminal state $s \in S$, with probabilities given by: $P_F(\tau) = P_F(s_0 \to \ldots \to s_n) = \prod_{t=0}^{n-1} P_F(s_{t+1}|s_t) \quad \forall \tau \in \mathcal{T}$. The distribution over complete trajectories that arises from a forward policy satisfies a Markov property. The forward policy can then be used to sample terminal states $x \in X$ by starting at state $s_0$ and iteratively sampling actions from $P_F$. 
A backward policy $P_B(\tau) = P_B(s_n \to \ldots \to s_0) = \prod_{t=0}^{n-1} P_B(s_{t}|s_{t+1}) \quad \forall \tau \in \mathcal{T}$. If $F$ is markovian flow, then $P_F$ and $P_B$ can be computed in terms of state and edge flows as: $P_F(s'|s) = \frac{F(s \to s')}{F(s)}$ and $P_B(s|s') = \frac{F(s \to s')}{F(s')}$.

\noindent Given a non-negative reward function $R : \mathcal{X} \rightarrow \mathbb{R}_{\geq 0}$, GFlowNets aim to learn a policy such that the probability of sampling a state $x \in \mathcal{X}$ is proportional to $R(x)$. The marginal likelihood of sampling a state $x \in X$ is the sum of likelihoods of all complete trajectories that terminate at $x$. If the objective function is globally minimized, then the likelihood of terminating at state $x$ is proportional to $R(x)$. Formally, the learning problem solved by GFlowNets is to estimate a policy $P_F$ over trajectories such that there exists a normalizing constant $Z$ satisfying: $R(x) = Z \sum_{\tau=(s_0 \rightarrow \ldots \rightarrow s_n = x)} P_F(\tau)\quad  \forall x \in \mathcal{X}$, where $Z= F(s_0) = \sum_{\tau \in \mathcal{T}}  F(\tau)$ is total flow at the initial state, and $\tau \in \mathcal{T}$ is the trajectory.

\subsection{Motivating Example}
\label{app:motivating_ex}

\begin{figure*}
    \centering
    \includegraphics[width=0.8\linewidth]{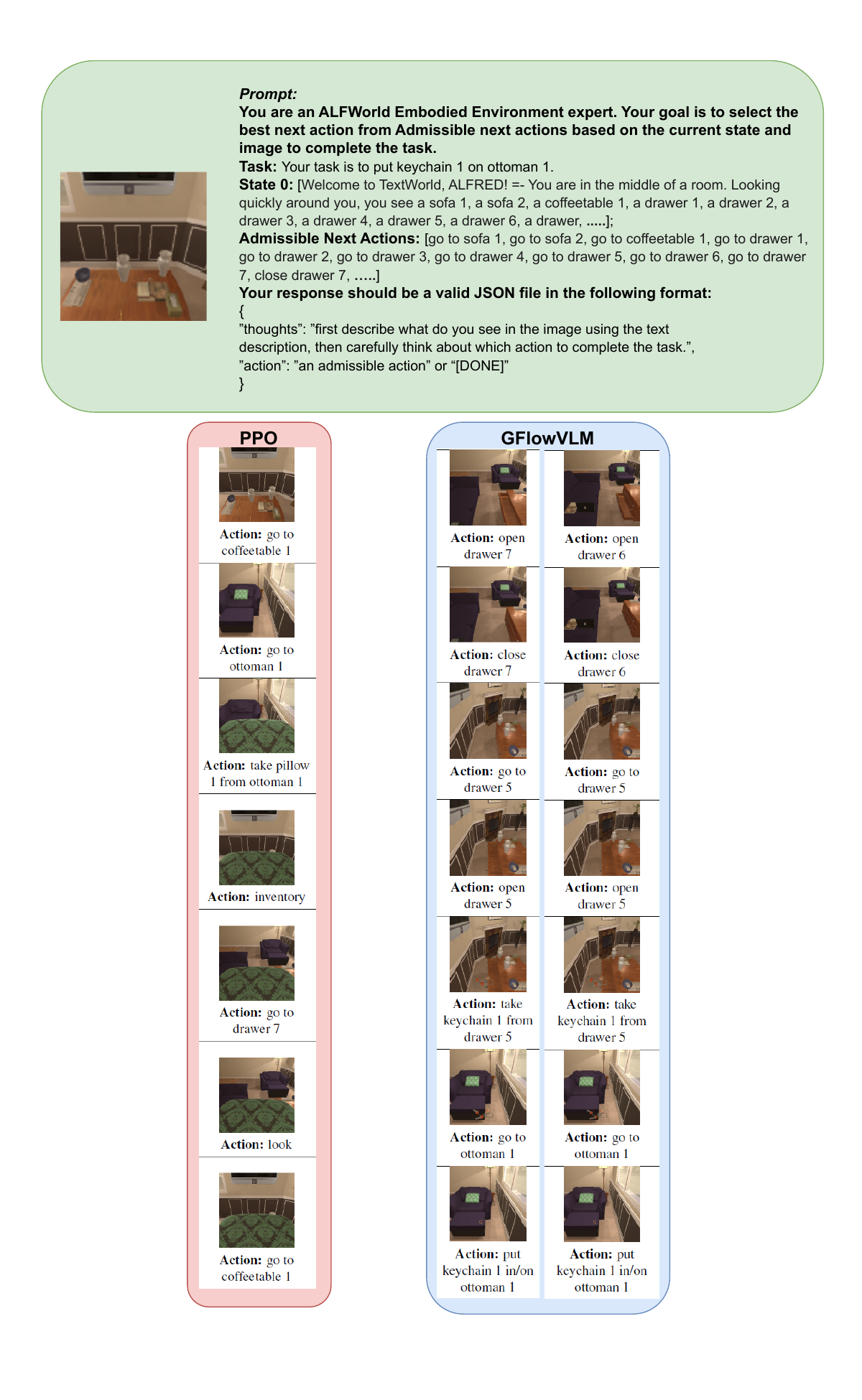}
    \caption{Overview of the prediction of diverse sequence using GFlowVLMs as compared to PPO for AlfWorld scenarios. The model takes the image of sequence and prompt as input, and generates the next number of sequence by implicitly modeling the causality.}
    \label{fig:teaser_alfworld}
\end{figure*}

We include a practical example from ALFWorld demonstrating how GFlowVLM can be applied to embodied AI tasks in~\cref{fig:teaser_alfworld}. The agent is presented with a visual observation of a simulated household environment and a high-level goal in natural language, such as "Put keychain in ottoman." The task requires the agent to generate a valid sequence of actions (e.g., open drawer $\rightarrow$ take keychain from drawer $\rightarrow$ close the opened drawer $\rightarrow$ go to ottomann $\rightarrow$ place keychain in ottoman). Importantly, there are multiple valid plans that achieve the same goal, with subtle causal constraints (e.g., the drawer must be open before taking the keychain from it, and objects must be picked up before being moved). We observe that models trained with PPO tend to converge on the most common or shortest path, while GFlowVLM generates a more diverse set of valid action sequences, reflecting a richer understanding of the causal structure of the environment. This example demonstrates GFlowNets’ strength in reasoning over multimodal inputs and learning structured, stochastic policies that preserve functional diversity. 

\section{Environments}
\subsection{ALFWorld}
\label{sec:alfworld_supp}
ALFWorld~\cite{shridhar2020alfworld} is an embodied AI environment combining a text-based interactive setup with a vision-language planning dataset. It includes six goal-conditioned tasks: ``Pick \& Place'', ``Examine in Light'', ``Clean \& Place'', ``Heat \& Place'', ``Cool \& Place'', and ``Pick Two \& Place''. The agent must plan and act based on visual cues and textual instructions (e.g., “go to shelf 1”) that specify the task. Unlike gym\_cards, where all states share the same action space, ALFWorld has a state-dependent action space $\mathcal{A}_t$; actions are context-dependent (e.g., ``put some pillows on the armchair'', the agent can only place a pillow after picking it up). Our prompt instructs the VLM to choose from the admissible actions $\mathcal{A}_t$, and we evaluate Out-Of-Distribution (OOD) performance using a test set of previously unseen scenes (see detailed prompt templates in ~\cref{tab:prompt-alf-cot-wohistory} and ~\cref{tab:prompt-alf-cot-history}). We use the same non-negative components of the reward function used in \cite{zhai2024fine}, which includes sub-goal and goal rewards: $r(s_t, a_t, s_{t+1}|g_{task}) = 50*1\{s_{t+1}=g_{task}\} + 1\{s_{t+1}=g_{task}\}$. We do not include the negative component of the reward function represented as $-1\{a_t \notin \mathcal{A}_{t}(s_t)\}$ in \cite{zhai2024fine}, since the actions are always selected from the admissible actions provided in the input prompt $p_t$. The rewards are non-negative, making it suitable for Var-TB and SubTB losses. However, since it lacks dense rewards for every transition, we didn't use GFlowVLM with DB loss on this task. 

\subsection{NumberLine}
\label{sec:numberline_supp}
This task involves moving a number along a synthetic number line to reach a target. NumberLine requires identifying two numbers in an image: ``target: $c$'' and ``current: $y_t$'', where $c$ and $y_t$ are both integers such that $c, y_t \in [n_{min}, n_{max}]$. The agent’s goal is to align $y_t$ with $c$ by outputting an action $a_t$ from the discrete set \{``+'', ``-'', \texttt{[DONE]}(if applicable)\}. Actions ``+'' and ``-'' adjust $y_t \pm 1$, while \texttt{[DONE]} signals task completion (see detailed prompt template in~\cref{tab:prompt-Numberline}). An episode ends when the $y_t = x$, or when the maximum step $T=2n_{max}$ is reached, which is the default setup of the environment. We set $n_{min}$ and $n_{max}$ as 0 and 5, respectively for the in-distribution examples, and set $n_{min}$ and $n_{max}$ as 10 and 50 for generating OOD examples. In the reward function used in \cite{zhai2024fine}, an agent receives a reward of $r(s_t, a_t) = 1$ when $y_{t+1}=c$, a penalty of $r(s_t, a_t) = -1$ upon taking an action that does not move the current number $y_t$ to the target $c$, and a reward of $r(s_t, a_t) = 0$, otherwise. For GFlowVLM, we revise the reward function with non-negative values as GFlowNets inherently require non-negative as follows:

\begin{equation}
R(x) = R(c, y_t) = \frac{l}{|c - y_t| + 1}
\end{equation}

where $l$ is a scaling constant set to 100.  This reward incentivizes the model to bring the current number closer to the target, progressively increasing the reward as the gap decreases. For fair comparison, we run RL4VLM~\cite{zhai2024fine} with revised reward structure. 

\subsection{Blackjack}
\label{sec:blackjack_supp}
The Blackjack task requires the VLM to reason with visual information and adapt to stochastic outcomes. The observation  $o_t$ includes two dealer cards (one face-down) and the player’s cards. The agent aims to win by selecting an action $a_t$ from \{``stand'', ``hit'', \texttt{[DONE]}(if applicable)\} (see detailed prompt template in ~\cref{tab:prompt-blackjack}). In the reward function used in \cite{zhai2024fine}, an agent receives a reward of $r(x) = 1, 0, -1$ upon win, draw and loss, respectively. Since GFlowNets inherently require non-negative reward, we revise the reward function to replace non-negative values as follows:

\begin{equation}
R(x) = \max(1 \times 10^{-10}, (r(x) + 1) \times 10),
\end{equation}
where $r(x)$ represents the environment's original reward for state $x$. This scales the rewards and ensures they are strictly non-negative. For fair comparison, we run RL4VLM~\cite{zhai2024fine} with revised reward structure. 

$R(x)$ represents the desirability or quality of a complete trajectory with final state $x$, similar to RL. It defines the target distribution from which the GFlowNet learns to sample, where higher-reward outcomes should be sampled more frequently.


\section{Training Objectives}
\label{app:loss_fn}
We adopt three different objective functions of GFlowNets, \textit{Trajectory-Balance (TB)}, \textit{Subtrajectory-Balance (SubTB)}, and \textit{Detailed-Balance (DB)}, to fine tune the VLM.

\subsection{Variance Trajectory Balanced (Var-TB) Loss}
\label{app:var-tb}
The \textit{Trajectory-Balanced} (TB) objective~\cite{malkin2022trajectory} ensures that the probability of generating a complete trajectory $\tau = (s_0 \rightarrow s_1 \rightarrow \cdots \rightarrow s_n = x)$ is proportional to the reward $R(\tau)$. Under the Markovian assumption, the forward policy $P_F(s_t|s_{t-1})$ transitions from state $s_{t-1}$ to $s_t$, while the backward policy $P_B(s_{t-1}|s_t)$ ensures consistency between forward and backward flows. This objective is given by:

\begin{flalign}
\begin{split}
Z\prod_{t=1}^n P_F(s_t|s_{t-1}; \theta) = R(x)\prod_{t=1}^n P_B(s_{t-1}|s_t; \theta),
\end{split}
\end{flalign}

where $Z$ is the partition function that normalizes the distribution. 

We now change $s$ to $z$ to match our definition of state in the main paper, where $z_{0:t}$ consists of a visual observation $o_t$ and an input prompt $p_t$ containing goal description, history states $s_{0:t-1}$, history actions $a_{0:t-1}$, and admissible actions $\mathcal{A}_{t}$. We use $\top$, which is the \texttt{[DONE]} symbol, to represent the terminal state $x$ of a trajectory. We adopt this notation because, in practice, the VLM predicts the action $\top$ to signify termination. This practical adaptation ensures consistency between the theoretical representation of terminal states and the actual predictions made by the VLM during inference.

Under the non-Markovian assumption of generating a complete trajectory $\tau = (z_0 \rightarrow z_1 \rightarrow \cdots \rightarrow z_n = x)$, and after adding goal into condition, we have:

\begin{flalign}
\begin{split}
Z\prod_{t=1}^n P_F(z_t |z_{0:t-1}, g;\theta) = R(x)\prod_{t=1}^n P_B(z_{t-1}|z_{t:n}, g ; \theta),
\end{split}
\end{flalign}

From \cite{zhang2023robust}, an estimation $Z$ for each trajectory $\tau$ can be expressed as:

\begin{flalign}
\label{est. init flow}
\begin{split}
\mathrm{\zeta}(\tau;\theta, g) = \log \frac {\prod_{t=1}^{n} P_F(z_t |z_{0:t-1}),g ; \theta)}{R(x) \prod_{t=1}^{n} P_B(z_{t-1}|z_{t:n},g ; \theta)} \\ 
= \log \frac {\prod_{t=1}^{n} P_F(z_t |z_{0:t-1},g ; \theta)}{R(x)}
\end{split}
\end{flalign}

where $P_B=1$ in our case since we formulate the trajectories as a tree structure, where a child state has only one parent state. In the optimal case, $\zeta(\tau;\theta, g)$ is equal to true $log Z$. The Variance-Trajectory-Balanced loss function aim to minimize the variance of $\mathrm{\zeta}(\tau;\theta, g)$ across trajectories to make the balance of the trajectories. The final Variance-Trajectory-Balanced loss is then defined as:

\begin{flalign}
\begin{split}
\mathcal{L}_{\mathrm{VarTB}}(\boldsymbol{\tau}; \theta) = \frac{1}{K} \sum_{k=1}^{K} \bigg(\mathrm{\zeta}(\tau_k; \theta, g) - \mathbb{E}_{\boldsymbol{\tau}}\big[\mathrm{\zeta}(\boldsymbol{\tau}; \theta, g)\big] \bigg)^2,
\end{split}
\end{flalign}


where $K$ represents the number of sampled trajectories. This loss ensures that high-reward trajectories are sampled more frequently by the policy.

\subsection{Subtrajectory Balanced (SubTB) Loss}
The \textit{Subtrajectory-Balanced} (SubTB) loss~\cite{madan2023learning} operates on subtrajectories of the form $\tau = (z_0 \rightarrow z_1 \rightarrow \cdots \rightarrow z_m)$. The subtrajectory balance ensures that each segment of the reasoning path or structure remains consistent, where the flows are balanced locally between forward and backward transitions. Under the non-Markovian assumption and after adding goal into conditions, the subtrajectory balance condition is expressed as:



\begin{flalign}
\begin{split}
F(z_0)\prod_{t=1}^m P_F(z_t |z_{0:t-1}),g ; \theta) = \\ F(z_m)\prod_{t=1}^m P_B(z_{t-1} |z_{t:m}),g ; \theta),
\end{split}
\end{flalign}

where $F(z_0)$ and $F(z_m)$ represent the flow into the initial ($z_0$) and final state ($z_m$) of the subtrajectory, respectively. Following \cite{pan2023better}, when all states $z_t$ are terminable with $\top$, we have $F(z_t)P_F (\top|z_{0:t}) = R(\top)$. Then the SubTB loss can be formulated as:
\begin{flalign}
\begin{split}
\mathcal{L}_{\mathrm{SubTB}}(z_{0:m},g ; \theta)=\sum_{0 \leq i < j \leq m} \\ \Bigg(\log \frac{R(z_{0:i}\top) \prod_{k=i+1}^{j} P_F(z_k | z_{0:k-1} ,g ; \theta) P_F(\top | z_{0:j} ,g ; \theta)}  {R(z_{0:j}\top)  P_F(\top | z_{0:i} ,g ; \theta)}\bigg)^2
\end{split}
\end{flalign}
where $\top$ is the  \texttt{[DONE]} symbol, denoting a trivial terminal state, and process continues until \texttt{[DONE]} symbol $\top$ is generated similar to \cite{hu2023amortizing}. This loss penalizes discrepancies in local transitions and ensures that all subsegments of a trajectory follow the correct balance conditions, reducing variance in smaller parts of the trajectory.

\subsection{Detailed Balanced (DB) Loss}


The \textit{Detailed-Balanced} (DB) loss~\cite{bengio2023gflownet} is used to ensure that each transition $s_t \rightarrow s_{t+1}$ between two states is balanced by matching the forward and backward flows at every step of the trajectory. The detailed balance condition is expressed as:

\begin{flalign}
\begin{split}
F(s_t) P_F(s_{t+1}|s_t) = F(s_{t+1}) P_B(s_t|s_{t+1}),
\end{split}
\end{flalign}

where $F(s_t)$ and $F(s_{t+1})$ represent the flow at states $s_t$ and $s_{t+1}$, respectively. Under the non-Markovian assumption of generating a complete trajectory $\tau = (z_0 \rightarrow z_1 \rightarrow \cdots \rightarrow z_n \rightarrow \top)$, where $\top$ is the terminal state of the sequence, DB loss is formulated as:

\begin{flalign}
\begin{split}
\mathcal{L}_{\mathrm{DB}}(z_{0:t} \rightarrow z_{0:t+1},g ; \theta) = \\ \Bigg(\log \frac{R(z_{0:t}\top) P_F(z_{t+1} | z_{0:t} ,g ; \theta) P_F(\top | z_{0:t+1} ,g ; \theta)}  {R(z_{0:t+1}\top)  P_F(\top | z_{0:t} ,g ; \theta)}\bigg)^2.
\end{split}
\end{flalign}

This loss ensures that every state-to-state transition follows the correct flow, preventing inconsistencies in the trajectory construction.

\paragraph{Comparisons of Loss Functions} TB loss controls the variance of $\zeta$ for the sampled trajectories, not the individual trajectory. Its main role is to bias sampling so that trajectory selection probability aligns with rewards~\cite{hu2023gflownet}. In addition, DB loss excels with dense rewards by ensuring flow consistency at each state, while SubTB and TB perform better in sparse settings by optimizing flow across (sub)trajectories. Additionally, TB is suited for tasks with known full sequences, and SubTB for costly large-trajectory sampling. 

\paragraph{Computational Complexity} In practice, we calculate (sub)trajectory or transition-based loss functions, which operate over (sub)trajectories or sampled transitions rather than the full state space. This allows us to efficiently handle the non-Markovian dependencies with \textit{linear} complexity. 


\section{Details of Experimental Setup}
In this section, we outline the experimental setup used to evaluate our approach across various tasks. We describe the key components of our implementation, including the data collection, diversity metric, and hyperparameters. By providing these details, we aim to ensure reproducibility and clarify how the proposed method integrates into different experimental frameworks.

\subsection{Off-Policy Data Collection}
\label{app_off_policy}
In this section, we describe our approach to off-policy data collection used in GFlowVLM for two distinct tasks, Numberline and Blackjack, emphasizing the integration of high-quality trajectories to enhance model training. These strategies ensure that the model learns from both successful and diverse trajectories, even when its on-policy performance falls short.

\begin{figure}
    \centering
    \includegraphics[width=\linewidth]{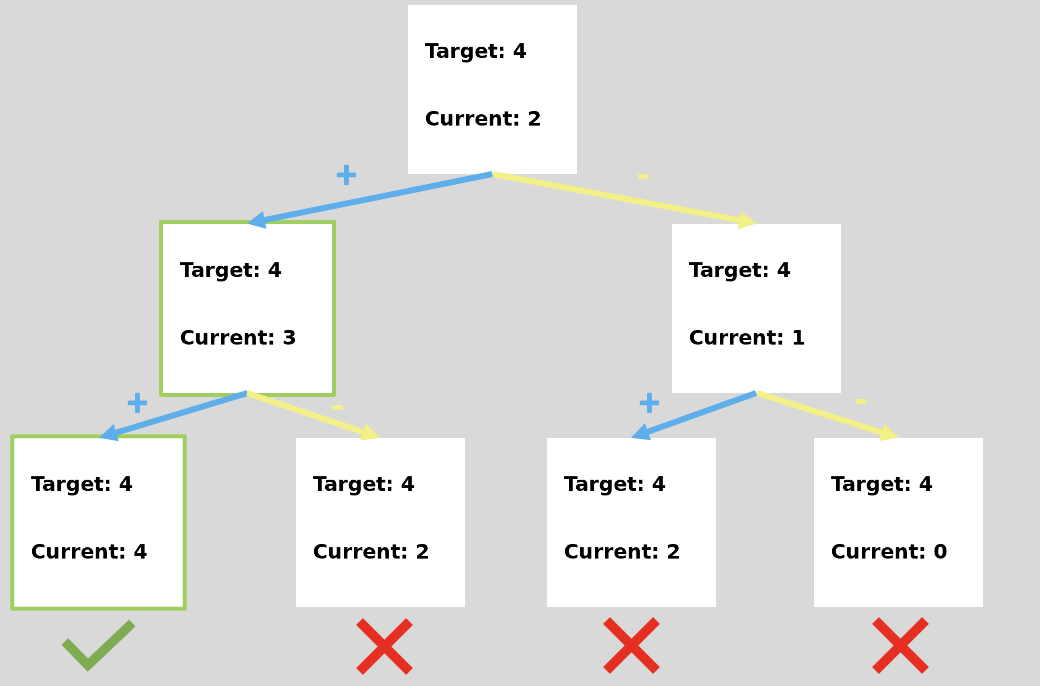}
    \caption{An example of off-policy data collection for NumberLine in a tree structure.}
    \label{fig:nl_fig}
\end{figure}

\paragraph{Numberline} During training, if the on-policy trajectory generated by the model fails to move the current number correctly towards the target, we augment the dataset by adding an off-policy, ground-truth trajectory to the buffer. These ground-truth trajectories represent successful paths that the model can follow to achieve the goal. By incorporating these accurate trajectories, we provide the model with additional supervision, which helps it learn to generalize better to unseen instances. This ensures the model benefits from examples of correct behavior, even when its predictions deviate from the optimal path. ~\cref{fig:nl_fig} illustrates the generation of both correct and incorrect trajectories, highlighting how diversity in training trajectories is encouraged to improve robustness.

\begin{figure}
    \centering
    \includegraphics[width=\linewidth]{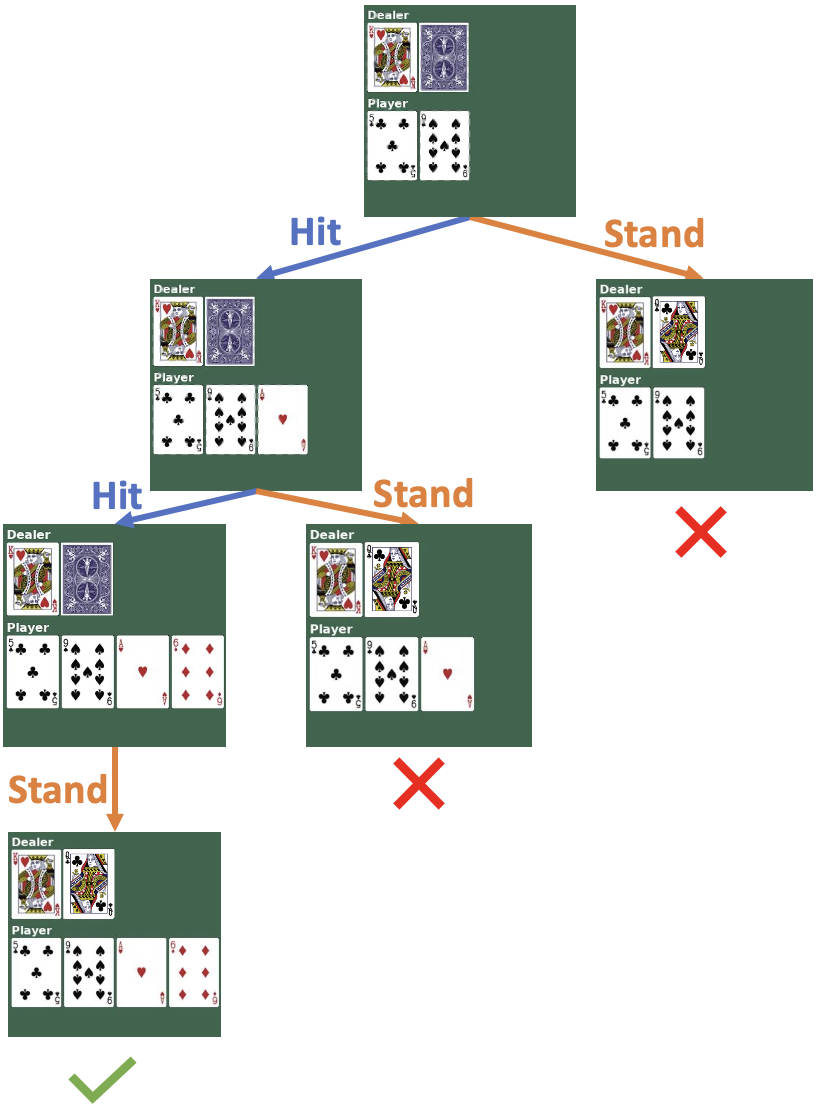}
    \caption{An example of off-policy data collection for BlackJack in a tree structure.}
    \label{fig:bj_fig}
\end{figure}

\paragraph{Blackjack} For the stochastic Blackjack task, deterministic ground-truth trajectories are not directly available due to the probabilistic outcomes of card draws. Instead, we generate high-quality off-policy trajectories using a \textit{rule-based heuristic}: The agent "stands" when the hand value is 17 or higher and "hits" otherwise. This strategy aligns with fundamental Blackjack principles, balancing the risk of exceeding a hand value of 21 against the potential for improvement by drawing additional cards. By leveraging this rule-based approach, we ensure that the training buffer includes trajectories that reflect a realistic yet principled decision-making process. Figure \ref{fig:bj_fig} demonstrates how both correct and incorrect trajectories are generated in a tree structure, promoting diversity in the training data and enabling the model to better handle a range of scenarios.

\subsection{SFT Dataset Collection}
To create the SFT dataset,  we iteratively interact with the environment to generate successful trajectories. For each successful trajectory, we manually append the ``\texttt{[DONE]}" token as the final action in the last state, explicitly marking the completion of the task. This approach aims to teach the model to predict the ``\texttt{[DONE]}" token as the appropriate action when the goal state is achieved.

\paragraph{Numberline} 
For the Numberline task, we execute ground-truth actions in the environment until the current state matches the target state. At this point, we append the ``\texttt{[DONE]}" token to indicate task completion. This process generated 8,000 data points with ``\texttt{[DONE]}" actions and 20,000 additional data points for other actions, using the base SFT dataset in \cite{zhai2024fine}.

\paragraph{Blackjack} 
For Blackjack, we adhere to the standard 17-point rule to determine actions. When the optimal decision is to take no further action, we append the ``\texttt{[DONE]}" token to the trajectory. This yielded 15,000 data points with ``\texttt{[DONE]}" actions and 50,000 for other actions, utilizing the SFT dataset from \cite{zhai2024fine}.

\paragraph{ALFWorld} 
For ALFWorld, we rely on expert actions derived from a heuristic \cite{shridhar2020alfworld}. At the end of each successful trajectory, we append the ``\texttt{[DONE]}" token to signify task completion. This resulted in 15,000 data points with ``\texttt{[DONE]}" actions and 45,000 for other actions using the SFT dataset from \cite{zhai2024fine}.

\subsection{Diversity Metric}

The diversity metric introduced in \cite{yu2024flow} calculates the diversity of successful trajectories found by a policy under the same number of samplings at inference time. Specifically, it is defined as follows:

\begin{equation}
\textit{Div} = \frac{\sum_{i=1}^n S_i \cdot \mathbbm{I}(S_i \geq 1)}{\sum_{i=1}^n \mathbbm{I}(S_i \geq 1)} \geq 1
\end{equation}

where \(n\) is the total number of tasks, \(S_i\) is the number of successful trajectories found for the \(i\)-th task, and \(\mathbbm{I}(S_i \geq 1)\) is an indicator function that equals 1 if at least one successful trajectory is found for the \(i\)-th task, and 0 otherwise. The denominator represents the number of tasks where the model finds at least one successful trajectory, while the numerator sums the total number of successful trajectories across all tasks. The smallest possible \textit{Div} is 1, indicating that a method finds at least one successful trajectory on average. For example, a \(\textit{Div} = 1.2\) suggests that, on average, a method finds 1.2 different successful trajectories. The (\textit{Div}@N) metric used in the main paper represents the diversity of successful trajectories after sampling $N$ trajectories' samples. 

\subsection{General Setup for Baselines and GFlowVLM}

All experiments are conducted on an H100 DGX machine with 80GB of memory. During VLM training, we directly optimize all trainable components, including the vision encoder, LLM, and MLP projector. For baseline methods, we utilize the open-source implementations provided in the original papers for SFT and RL4VLM~\cite{zhai2024fine}. A \textit{CosineAnnealingLR} scheduler is adopted, starting with an initial learning rate of $1 \times 10^{-5}$, decaying to a final learning rate of $1 \times 10^{-9}$, and reaching its maximum learning rate at step 25. For GFlowVLM, a buffer size of 4 is used across all tasks. To ensure a fair comparison, we report the number of environment steps for each method.

\subsection{CoT Weighting Factor $\lambda$}
\label{section-cot-ablation}

\begin{figure}
    \centering
    \includegraphics[width=0.80\linewidth]{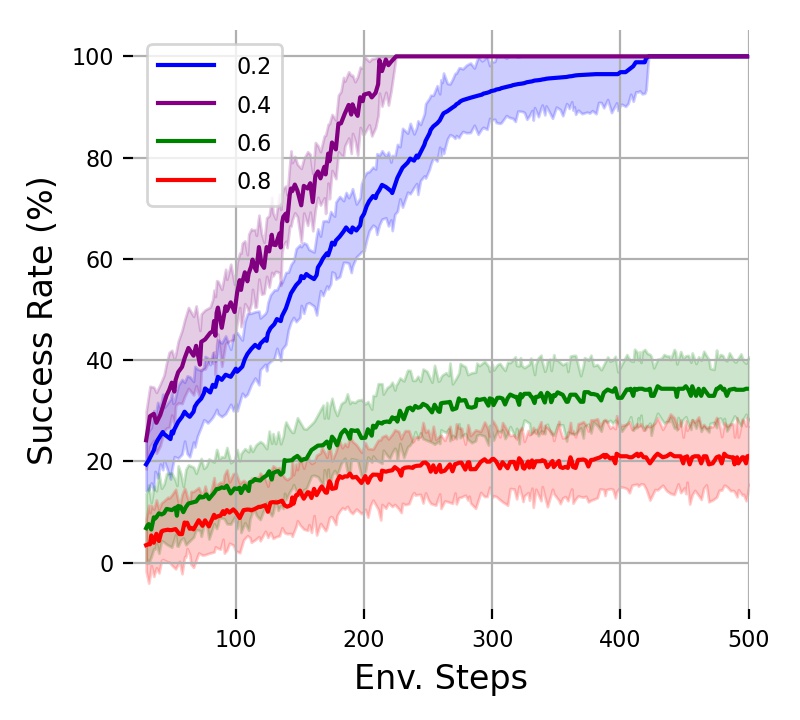}
    \caption{Average success rates (\%) of our method under different CoT weighting factor $\lambda$ on NumberLine across three loss functions. }
    \label{fig:cot-ablation}
\end{figure}

\begin{flalign}
\begin{split}
P_F(z_{t+1}| z_{0:t}, g; \theta) = P_{\text{Action}}(a_t | z_{0:t},  c_t, g; \theta) + \\
\lambda P_{\text{CoT}}(c_t | z_{0:t},  g; \theta),
\end{split}
\label{cot-factor-equ-supp}
\end{flalign}

The CoT weighting factor, $\lambda \in [0, 1]$, controls the influence of CoT reasoning within our framework, as discussed briefly in the main paper (rewritten here in ~\cref{cot-factor-equ-supp}). To assess the impact of $\lambda$, we compute the average performance of our proposed framework, GFlowVLM, using three loss functions, each evaluated with four random seeds. As shown in Figure \ref{fig:cot-ablation}, a moderate  $\lambda$ (e.g., 0.4) yields the best performance on NumberLine tasks across three different loss functions.  When $\lambda$ is too high (0.8) or too low (0.2), $P_{\text{CoT}}(c_t | z_{0:t}, g; \theta)$ or $P_{\text{Action}}(a_t | z_{0:t}, c_t, g; \theta)$ overly influences the estimation of $P_F$, respectively, leading to imbalanced learning dynamics. Thus, setting $\lambda=0.4$ effectively balances CoT and action learning, enhancing reasoning performance. We use the same value of $\lambda=0.4$ across all experiments in this work.



\begin{table*}[th!]
\small
\centering
\begin{tabular}{lcccccc}
\toprule
\textbf{Method} & \textbf{Train Data} & \textbf{Assump.} & \textbf{SFT Init.} & \textbf{NL} & \textbf{NL-OOD} & \textbf{BJ} \\ 
\midrule
\hline
\multicolumn{6}{c}{\textbf{Ablations of RL4VLM}}\\
RL4VLM~\cite{zhai2024fine}$^*$ & On & M & \checkmark & 34.8 & 1.9 & 23.5 \\ 
RL4VLM~\cite{zhai2024fine} & On & M  & \checkmark & 89.4 & 3.1 & 40.2 \\
RL4VLM~\cite{zhai2024fine} & On & NM  & \checkmark & \textbf{90.3} & \textbf{4.4} & \textbf{41.0} \\ 
\midrule
 \multicolumn{6}{c}{\textbf{Ablations of GFlowVLM w/ Var-TB w/ On and Off-Policy}} \\
GFlowVLM w/ Var-TB  & On & M & \checkmark & 93.4 & 4.7 & 41.0\\ 
GFlowVLM w/ Var-TB  & On & NM & \checkmark & \textbf{100.0} & 6.2 & 41.4 \\ 
GFlowVLM w/ Var-TB & Off & M & \checkmark &  94.5 & 17.2 & 42.0 \\ 
GFlowVLM w/ Var-TB & Off & NM & \checkmark & \textbf{100.0} & \textbf{17.3} & \textbf{43.0} \\ 
\midrule
 \multicolumn{6}{c}{\textbf{Ablations of GFlowVLM w/ SubTB w/ On and Off-Policy}}\\
 GFlowVLM w/ SubTB & On & M & \checkmark & 91.7 & 4.0 & 40.2 \\ 
GFlowVLM w/ SubTB & On & NM & \checkmark& \textbf{100.0} & 7.0 & 41.7 \\ 
GFlowVLM w/ SubTB & Off & M & \checkmark &94.8 & \textbf{17.3} & 40.5  \\ 
GFlowVLM w/ SubTB & Off & NM & \checkmark & \textbf{100.0} & 16.7 & \textbf{42.4} \\ 
\midrule
\multicolumn{6}{c}{\textbf{Ablations of GFlowVLM w/ DB w/ On and Off-Policy}}\\
 
GFlowVLM w/ DB & On & M & \checkmark &90.1 & 5.3 & 40.0 \\
GFlowVLM w/ DB & On & NM & \checkmark& \textbf{100.0} & 9.1 & 42.2 \\ 
GFlowVLM w/ DB & Off & M & \checkmark & 93.6 & 16.3 & 41.5 \\ 
GFlowVLM w/ DB & Off & NM & \checkmark & \textbf{100.0} & \textbf{18.6} & \textbf{43.8} \\ 
\bottomrule
\end{tabular}
\caption{Ablations of GFlowVLM with Markovian assumption for NumberLine (NL) and BlackJack (BJ) tasks for in-distribution and out-of-distributions (OOD) tasks. $^*$We use the same reward function as ours. NL-OOD stands for Number line with out-of-distribution tasks. On and Off represent On-Policy and Off-Policy, respectively. M and NM stands for Markovian and non-Markovian assumption respectively.}
\label{tab:markovian-results-nl-bj}
\end{table*}

\begin{table*}[ht!]
\small
\centering
\begin{tabular}{lcccccccccc}
\toprule
\textbf{Method} & \textbf{Assump.} & \textbf{Pick} & \textbf{Look} & \textbf{Clean} & \textbf{Heat} & \textbf{Cool} & \textbf{Pick2} & \textbf{Avg. } & \textbf{OOD} & \textbf{Div@16} \\ 
\midrule
\hline
\multicolumn{11}{c}{\textbf{Ablations of RL4VLM}}\\
RL4VLM~\cite{zhai2024fine} & M  & 47.4 & \textbf{14.7} & \textbf{10.4} & 14.4  & 18.8  & 18.0 & 21.7 & 4.8  & \textbf{1.12} \\ 
RL4VLM~\cite{zhai2024fine} & NM & \textbf{49.1} & 13.5 & 9.8 & \textbf{15.2} & \textbf{20.1} & \textbf{20.6} & \textbf{22.1} &  \textbf{6.1} & 1.11 \\ 
\midrule
\multicolumn{11}{c}{\textbf{Ablations of GFlowVLM w/  SubTB}}\\
GFlowVLM w/ SubTB  & M  & 46.0 & 10.1 & 9.7 & 14.7  & \textbf{24.6}  & 23.7 & 22.1 & 8.0 & 1.34 \\ 
GFlowVLM w/ SubTB  & NM  & \textbf{50.0} & \textbf{23.1} & \textbf{10.0} & \textbf{18.7}  & 24.3  & 23.7 & \textbf{26.1} & \textbf{12.3} & \textbf{1.40} \\ 

\midrule
\multicolumn{11}{c}{\textbf{Ablations of GFlowVLM w/  Var-TB}}\\
GFlowVLM w/ Var-TB   & M  & 45.1 & 12.2 & \textbf{11.3} & \textbf{15.7}  & 20.6  & \textbf{24.7} & 22.9 & 7.6 & 1.37 \\ 
GFlowVLM w/ Var-TB  & NM  & \textbf{50.0} & \textbf{22.2} & 10.2 & 16.1  & \textbf{22.7}  & 21.9 & \textbf{25.7} & \textbf{10.9} & \textbf{1.41} \\ 
\bottomrule
\end{tabular}
\caption{Ablations of GFlowVLM with Markovian assumption for ALFWorld. Since Alfworld does not provide dense rewards, we can not not using DB loss here. Furthermore, while RL4VLM and GFlowVLM with SubTB are trained with SFT initialization, GFVLM with TB-Var is without STF initialization since we do not need to model the flow. M and NM stands for Markovian and non-Markovian assumption respectively.}
\label{tab:markovian-results-alfworld}
\end{table*}

\begin{table*}[t] 
\small
\renewcommand{\arraystretch}{1.5}
\centering
\begin{tabular}{|p{0.95\textwidth}|}
\toprule
\textbf{Image input}: 
\begin{center}
\fbox{\includegraphics[width=0.25\textwidth]{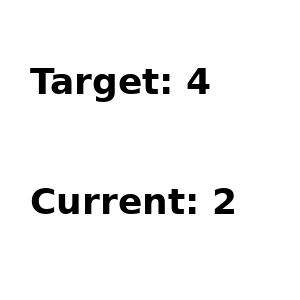}} 
\end{center}
\\

\midrule

\textbf{NumberLine prompt template without history information (Markovian)} \\ \hline
You are playing a game called number line. You will see a target number and a current number in the image. And your goal is to move the current number closer to the target by choosing either adding or subtracting one to the current number. You need to first give the thoughts and then you can choose between \texttt{["+", "-"]}. \textcolor{brown}{Use ``\texttt{[DONE]}" when you think you have completed the task.} Your response should be a valid JSON file in the following format: \\ 
\texttt{\{ } \\
\hspace*{1em} \textcolor{RedOrange}{``current number": "x",} \\
\hspace*{1em} \textcolor{RedOrange}{``target number": "x",} \\
\hspace*{1em} \textcolor{PineGreen}{``thoughts": ``{first read out the current and target number, then think }}  \\
\hspace*{1em} \textcolor{PineGreen}{ carefully about which action to choose",} \\
\hspace*{1em} \textcolor{purple}{``action": ``-" or ``+"} \textcolor{brown}{ or ``\texttt{[DONE]}"}  \newline
\texttt{\} } \\ \hline

\textbf{NumberLine prompt template with history information (non-Markovian)} \\ \hline
You are playing a game called number line. You will see a target number and a current number in the image. And your goal is to move the current number closer to the target by choosing either adding or subtracting one to the current number. Below are the history actions and states you've done. \\

State 0: 1 \\
Action 1: \texttt{"+"} \\
State 1: 2 \\

Based on the history information, you need to first give the thoughts and then you can choose between \texttt{["+", "-"]}. \textcolor{brown}{Use ``\texttt{[DONE]}" when you think you have completed the task.}Your response should be a valid JSON file in the following format: \\ 
\texttt{\{ } \\
\hspace*{1em} \textcolor{RedOrange}{``current number": "x",} \\
\hspace*{1em} \textcolor{RedOrange}{``target number": "x",} \\
\hspace*{1em} \textcolor{PineGreen}{``thoughts": ``{first read out the current and target number, then think }}  \\
\hspace*{1em} \textcolor{PineGreen}{carefully about which action to choose",} \\
\hspace*{1em} \textcolor{purple}{``action": ``-" or ``+"}\textcolor{brown}{ or ``\texttt{[DONE]}"}  \newline
\texttt{\} } \\ \hline

\end{tabular}
\caption{Prompt Template with Markovian and non-Markovian assump. for NumberLine. The sentence in \textcolor{brown}{brown} is only applicable for SubTB and DB losses.}
\label{tab:prompt-Numberline}
\end{table*}




\renewcommand{\arraystretch}{1.5} 

\begin{table*}[t] 
\renewcommand{\arraystretch}{1.5}
\centering
\begin{tabular}{|p{0.95\textwidth}|}
\toprule
\textbf{Image input}: 
\begin{center}
\includegraphics[width=0.25\textwidth]{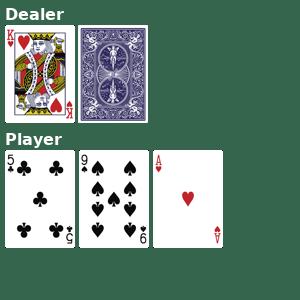}
\end{center}
\\
\midrule
\textbf{BlackJack prompt template without history information (Markovian)} \\ \hline
You are a blackjack player. You are observing the current game state. You need to first give an explanation and then you can choose between \texttt{["stand", "hit"]}. \textcolor{brown}{Use ``\texttt{[DONE]}" when you think you have completed the task.} Your response should be a valid JSON file in the following format: \\ 
\texttt{\{ } \\
\hspace*{1em} \textcolor{PineGreen}{``thoughts": ``{first describe your total points and the dealer’s total points then think about which action to choose}",} \newline
\hspace*{1em}  \textcolor{purple}{``action": ``stand" or "hit"} \textcolor{brown}{ or ``\texttt{[DONE]}"} \newline
\texttt{\} } \\ \hline

\textbf{BlackJack prompt template with history information (non-Markovian)} \\ \hline
You are a blackjack player. You are observing the current game state. Below are the history actions and states. \\

State 0: 14 points \\
Action 1: \texttt{"hit"} \\
State 1: 15 points \\

Based on the history information, you need to first give an explanation and then you can choose between \texttt{[``stand", ``hit"]}. \textcolor{brown}{Use ``\texttt{[DONE]}" when you think you have completed the task.} Your response should be a valid JSON file in the following format: \\ 
\texttt{\{ } \\
\hspace*{1em} \textcolor{PineGreen}{``thoughts": "{first describe your total points and the dealer’s total points then think about which action to choose}",} \newline
\hspace*{1em}  \textcolor{purple}{``action": ``stand" or "hit"} \textcolor{brown}{ or ``\texttt{[DONE]}"} \newline
\texttt{\} } \\ \hline

\end{tabular}
\caption{Prompt Templates with Markovian and non-Markovian assump. for BlackJack. The sentence in \textcolor{brown}{brown} is only applicable for SubTB and DB losses.}
\label{tab:prompt-blackjack}
\end{table*}

\begin{table*}[t] 
\renewcommand{\arraystretch}{1.5}
\centering
\begin{tabular}{|p{0.95\textwidth}|}
\toprule
\textbf{Image input}: 
\begin{center}
\includegraphics[width=0.25\textwidth]{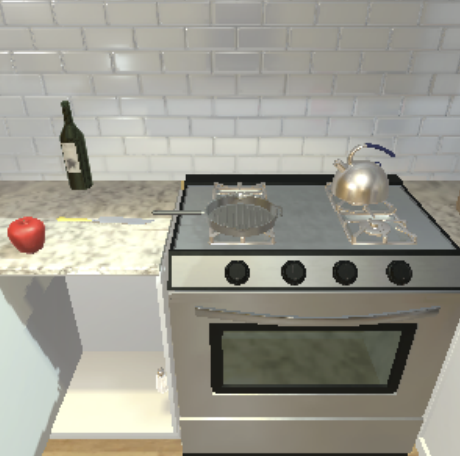}
\end{center}
\\
\midrule
\textbf{ALFWorld prompt template without history information  (Markovian)} \\ \hline

You are an ALFWorld Embodied Environment expert. Your goal is to select the best next action from the Admissible Next Actions based on the current state and image to complete the task. \textcolor{brown}{Use ``\texttt{[DONE]}" when you think you have completed the task.}\\

Task: Your task is to put a cool mug in cabinet. \\

Current State: \texttt{"['You arrive at loc 1. The cabinet 1 is open. On the cabinet 1, you see a pan 1, a kettle 1, a winebottle 1, a apple 1, a stoveknob 1, a stoveknob 2, a stoveknob 3, a stoveknob 4, a knife 1, a saltshaker 1, and a bread 1.']."} \\

Admissible Next Actions: \texttt{['go to countertop 1', 'go to cabinet 2', 'go to countertop 2', 'go to stoveburner 1', 'go to drawer 1', 'go to drawer 2', 'go to drawer 3', 'go to stoveburner 2', 'go to stoveburner 3', 'go to stoveburner 4', 'go to drawer 4', 'go to cabinet 3', 'go to cabinet 4', 'go to microwave 1', 'go to cabinet 5', 'go to cabinet 6', 'go to cabinet 7', 'go to sink 1', 'go to sinkbasin 1', 'go to fridge 1', 'go to toaster 1', 'go to coffeemachine 1', 'go to cabinet 8', 'go to drawer 5', 'go to drawer 6', 'go to drawer 7', 'go to drawer 8', 'go to shelf 1', 'go to shelf 2', 'go to countertop 3', 'go to shelf 3', 'go to drawer 9', 'go to garbagecan 1', 'open cabinet 1', 'close cabinet 1', 'take pan 1 from cabinet 1', 'take kettle 1 from cabinet 1', 'take winebottle 1 from cabinet 1', 'take apple 1 from cabinet 1', 'take stoveknob 1 from cabinet 1', 'take stoveknob 2 from cabinet 1', 'take stoveknob 3 from cabinet 1', 'take stoveknob 4 from cabinet 1', 'take knife 1 from cabinet 1', 'take saltshaker 1 from cabinet 1', 'take bread 1 from cabinet 1', 'inventory', 'look', 'examine cabinet 1']. } \\

Your response should be a valid JSON file in the following format: \\ 
\texttt{\{ } \\
\hspace*{1em} \textcolor{PineGreen}{``thoughts": "{first describe what do you see in the image using the text}} \newline
\hspace*{1em} \textcolor{PineGreen}{description, then carefully think about which action to complete the task.",} \\
\hspace*{1em} \textcolor{purple}{``action": ``an admissible action"} \textcolor{brown}{ or ``\texttt{[DONE]}"}\newline
\texttt{\} } \\ \hline
\end{tabular}
\caption{Prompt template with Markovian assump. for ALFWorld. The sentence in \textcolor{brown}{brown} is only applicable for SubTB and DB losses.}
\label{tab:prompt-alf-cot-wohistory}
\end{table*}

\begin{table*}[t] 
\small
\renewcommand{\arraystretch}{1.5}
\centering
\begin{tabular}{|p{0.95\textwidth}|}
\toprule
\textbf{Image input}: 
\begin{center}
\includegraphics[width=0.25\textwidth]{alf-example.jpeg}
\end{center}
\\
\midrule
\textbf{ALFWorld prompt template with history information (non-Markovian)} \\ \hline

You are an ALFWorld Embodied Environment expert. Your goal is to select the best next action from the Admissible Next Actions based on the previous and current states and image to complete the task. Use "\texttt{[DONE]}" when you think you have completed the task. \\

Task: Your task is to put a cool mug in cabinet. \\

State 0: \texttt{['-= Welcome to TextWorld, ALFRED! =- You are in the middle of a room. Looking quickly around you, you see a countertop 1, a coffeemachine 1, a cabinet 1, a cabinet 2, a cabinet 3, a sink 1, a cabinet 4, a drawer 1, a drawer 2, a drawer 3, a sinkbasin 1, a cabinet 5, a toaster 1, a fridge 1, a cabinet 6, a cabinet 7, a cabinet 8, a microwave 1, a cabinet 9, a cabinet 10, a cabinet 11, a drawer 4, a cabinet 12, a drawer 5, a stoveburner 1, and a stoveburner 2.']} 

Action 1: \texttt{"open cabinet 1."} 

State 1: \texttt{"['You arrive at loc 1. The cabinet 1 is open. On the cabinet 1, you see a pan 1, a kettle 1, a winebottle 1, a apple 1, a stoveknob 1, a stoveknob 2, a stoveknob 3, a stoveknob 4, a knife 1, a saltshaker 1, and a bread 1.']."} \\

Admissible Next Actions: \texttt{['go to countertop 1', 'go to cabinet 2', 'go to countertop 2', 'go to stoveburner 1', 'go to drawer 1', 'go to drawer 2', 'go to drawer 3', 'go to stoveburner 2', 'go to stoveburner 3', 'go to stoveburner 4', 'go to drawer 4', 'go to cabinet 3', 'go to cabinet 4', 'go to microwave 1', 'go to cabinet 5', 'go to cabinet 6', 'go to cabinet 7', 'go to sink 1', 'go to sinkbasin 1', 'go to fridge 1', 'go to toaster 1', 'go to coffeemachine 1', 'go to cabinet 8', 'go to drawer 5', 'go to drawer 6', 'go to drawer 7', 'go to drawer 8', 'go to shelf 1', 'go to shelf 2', 'go to countertop 3', 'go to shelf 3', 'go to drawer 9', 'go to garbagecan 1', 'open cabinet 1', 'close cabinet 1', 'take pan 1 from cabinet 1', 'take kettle 1 from cabinet 1', 'take winebottle 1 from cabinet 1', 'take apple 1 from cabinet 1', 'take stoveknob 1 from cabinet 1', 'take stoveknob 2 from cabinet 1', 'take stoveknob 3 from cabinet 1', 'take stoveknob 4 from cabinet 1', 'take knife 1 from cabinet 1', 'take saltshaker 1 from cabinet 1', 'take bread 1 from cabinet 1', 'inventory', 'look', 'examine cabinet 1']. } \\

Your response should be a valid JSON file in the following format: \\ 
\texttt{\{ } \\
\hspace*{1em} \textcolor{PineGreen}{"thoughts": "{first describe what do you see in the image using the text}} \newline
\hspace*{1em} \textcolor{PineGreen}{description, then carefully think about which action to complete the task.",} \\
\hspace*{1em} \textcolor{purple}{"action": "an admissible action"} \textcolor{brown}{ or ``\texttt{[DONE]}"}
\newline
\texttt{\} } \\ \hline
\end{tabular}
\caption{Prompt template with non-Markovian assump. for ALFWorld. The sentence in \textcolor{brown}{brown} is only applicable for SubTB and DB losses.}
\label{tab:prompt-alf-cot-history}
\end{table*}

\section{Qualitative Results}
\label{app:qual-results}
We present an example in ALFWorld in \cref{tab:alf-qual}, with the goal of "put some keychains on the ottoman" to illustrate key insights into our method.

Our method encourages exploration by sampling proportional to the reward, allowing it to avoid getting stuck in suboptimal states—a common limitation observed in PPO. This exploration not only prevents suboptimal convergence but also enables the model to generate more diverse solutions, as demonstrated by the multiple trajectories shown in ~\cref{tab:alf-qual}. Through repeated sampling, our method effectively considers a wider range of potential paths to achieve the goal.

PPO, in contrast, tends to rely on superficial semantic patterns to make decisions. For instance, it may prioritize reaching the "ottoman" directly without first retrieving the keychains, as the term "ottoman" semantically aligns with the goal. This behavior highlights the risk of overfitting to pattern recognition rather than aligning actions with the ultimate reward.



\begin{table}[htbp]
\small
    \centering
    \begin{tabular}{|c|c|c|}
    \hline 
    \multicolumn{3}{|c|}{\textbf{Goal: put some keychains on ottoman}.} \\
    \midrule
    \textbf{PPO} &\textbf{Ours-Traj. 1} & \textbf{Ours-Traj. 2} \\
    \hline
    \begin{minipage}[b]{0.30\linewidth}
        \centering
        \includegraphics[width=0.7\linewidth]{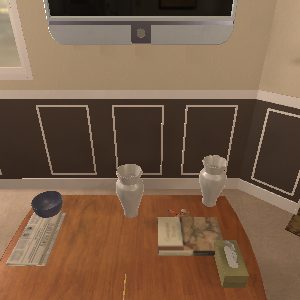} \\ 
        \textbf{Action:} go to coffeetable 1
    \end{minipage} &
    \begin{minipage}[b]{0.30\linewidth}
        \centering
        \includegraphics[width=0.7\linewidth]{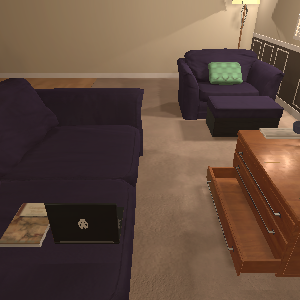} \\ 
        \textbf{Action:} open drawer 6
    \end{minipage} &
    \begin{minipage}[b]{0.30\linewidth}
        \centering
        \includegraphics[width=0.7\linewidth]{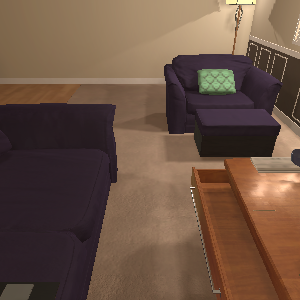} \\ 
        \textbf{Action:} open drawer 7
    \end{minipage} \\
    \hline

    \begin{minipage}[b]{0.30\linewidth}
        \centering
        \includegraphics[width=0.7\linewidth]{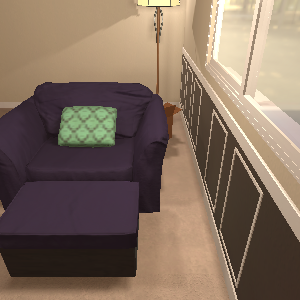} \\ 
        \textbf{Action:} go to ottoman 1
    \end{minipage} &
    \begin{minipage}[b]{0.30\linewidth}
        \centering
        \includegraphics[width=0.7\linewidth]{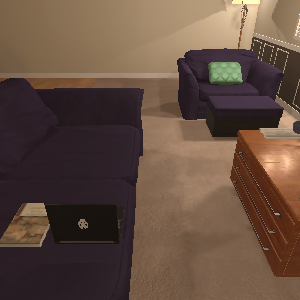} \\ 
        \textbf{Action:} close drawer 6
    \end{minipage} &
    \begin{minipage}[b]{0.30\linewidth}
        \centering
        \includegraphics[width=0.7\linewidth]{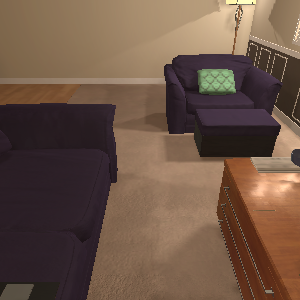} \\ 
        \textbf{Action:} close drawer 7
    \end{minipage} \\
    \hline
    \begin{minipage}[b]{0.30\linewidth}
        \centering
        \includegraphics[width=0.7\linewidth]{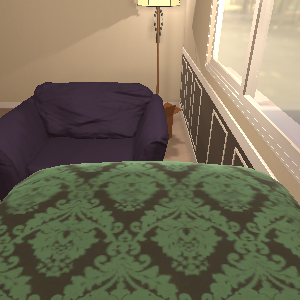} \\ 
        \textbf{Action:} take pillow 1 from ottoman 1
    \end{minipage} &
    \begin{minipage}[b]{0.30\linewidth}
        \centering
        \includegraphics[width=0.7\linewidth]{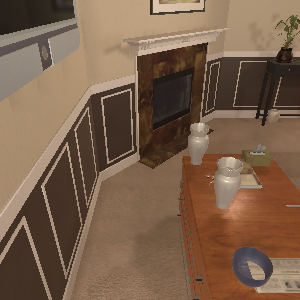} \\ 
        \textbf{Action:} go to drawer 5
    \end{minipage} &
    \begin{minipage}[b]{0.30\linewidth}
        \centering
        \includegraphics[width=0.7\linewidth]{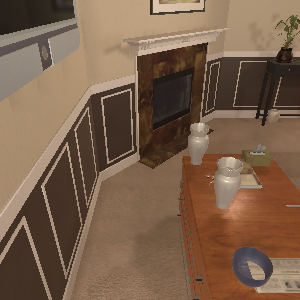} \\ 
        \textbf{Action:} go to drawer 5
    \end{minipage} \\
    \hline

    \begin{minipage}[b]{0.30\linewidth}
        \centering
        \includegraphics[width=0.7\linewidth]{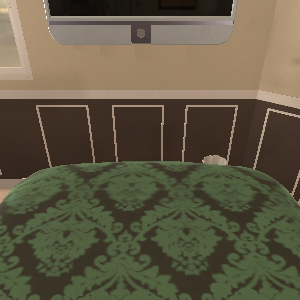} \\ 
        \textbf{Action:} inventory
    \end{minipage} &
    \begin{minipage}[b]{0.30\linewidth}
        \centering
        \includegraphics[width=0.7\linewidth]{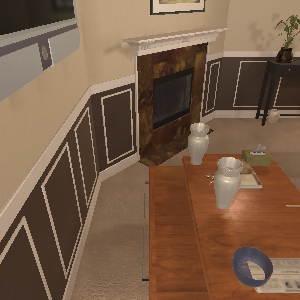} \\ 
        \textbf{Action:} open drawer 5
    \end{minipage} &
    \begin{minipage}[b]{0.30\linewidth}
        \centering
        \includegraphics[width=0.7\linewidth]{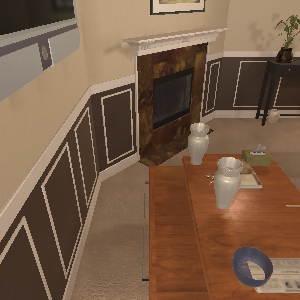} \\ 
        \textbf{Action:} open drawer 5
    \end{minipage} \\
    \hline

    \begin{minipage}[b]{0.30\linewidth}
        \centering
        \includegraphics[width=0.7\linewidth]{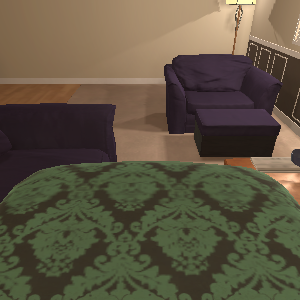} \\ 
        \textbf{Action:} go to drawer 7
    \end{minipage} &
    \begin{minipage}[b]{0.30\linewidth}
        \centering
        \includegraphics[width=0.7\linewidth]{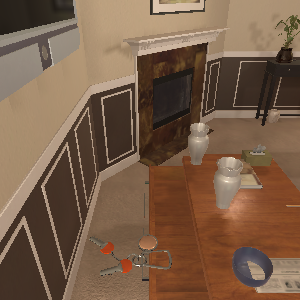} \\ 
        \textbf{Action:} take keychain 1 from drawer 5
    \end{minipage} &
    \begin{minipage}[b]{0.30\linewidth}
        \centering
        \includegraphics[width=0.7\linewidth]{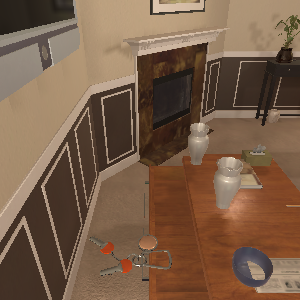} \\ 
        \textbf{Action:} take keychain 1 from drawer 5
    \end{minipage} \\
    \hline

    \begin{minipage}[b]{0.30\linewidth}
        \centering
        \includegraphics[width=0.7\linewidth]{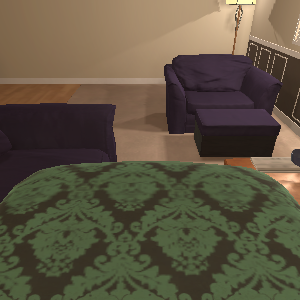} \\ 
        \textbf{Action:} look
    \end{minipage} &
    \begin{minipage}[b]{0.30\linewidth}
        \centering
        \includegraphics[width=0.7\linewidth]{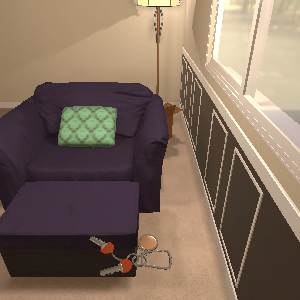} \\ 
        \textbf{Action:} go to ottoman 1
    \end{minipage} &
    \begin{minipage}[b]{0.30\linewidth}
        \centering
        \includegraphics[width=0.7\linewidth]{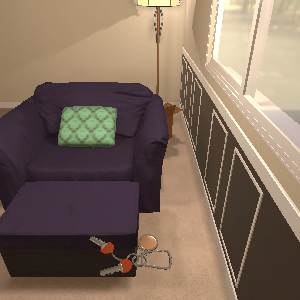} \\ 
        \textbf{Action:} go to ottoman 1
    \end{minipage} \\
    \hline

    \begin{minipage}[b]{0.30\linewidth}
        \centering
        \includegraphics[width=0.7\linewidth]{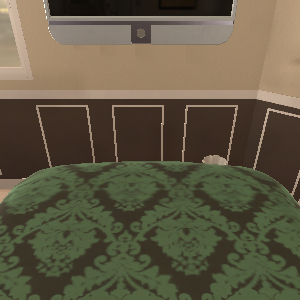} \\ 
        \textbf{Action:} go to coffeetable 1
    \end{minipage} &
    \begin{minipage}[b]{0.30\linewidth}
        \centering
        \includegraphics[width=0.7\linewidth]{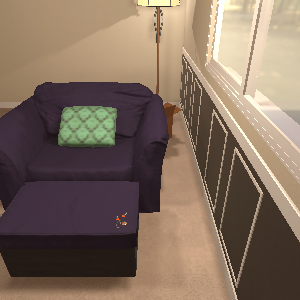} \\ 
        \textbf{Action:} put keychain 1 in/on ottoman 1
    \end{minipage} &
    \begin{minipage}[b]{0.30\linewidth}
        \centering
        \includegraphics[width=0.7\linewidth]{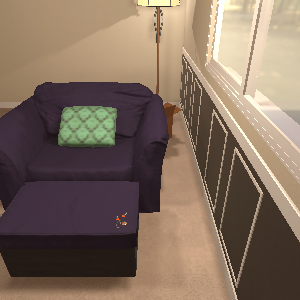} \\ 
        \textbf{Action:} put keychain 1 in/on ottoman 1
    \end{minipage} \\
    \bottomrule
    \end{tabular}
    \caption{Qualitative results for ALFWorld task. GFlowVLM generates diverse trajectories in contrast to PPO. }
    \label{tab:alf-qual}
\end{table}

\clearpage

\section{Ablation Study of Markovian and non-Markovian}

To evaluate the impact of Markovian and non-Markovian assumptions on performance, we conduct an ablation study with our method, GFlowVLM with both On-Policy and Off-Policy training, and RL4VLM \cite{zhai2024fine} across 3 tasks: NumberLine and Blackjack and ALFWorld. The primary difference between these two assumptions lies in the prompt template used during training. Under the Markovian assumption, the model operates with prompts that do not include historical information about prior actions and states, relying solely on the current state. Conversely, the non-Markovian assumption incorporates the history of actions and states into the prompt, providing richer contextual information (see prompt templates in~\cref{tab:prompt-Numberline},~\cref{tab:prompt-blackjack},~\cref{tab:prompt-alf-cot-history},~\cref{tab:prompt-alf-cot-wohistory} for details).

As shown in ~\cref{tab:markovian-results-nl-bj}, the non-Markovian assumption leads to consistently better performance across all tasks. In NumberLine and Blackjack, GFlowVLM achieves substantial improvements in both in-distribution and out-of-distribution scenarios under the non-Markovian assumption.For instance, in the Numberline task, GFlowVLM with the DB loss demonstrates improved out-of-distribution performance when transitioning from Markovian to non-Markovian assumptions. Specifically, with on-policy training, the performance increases from 5.3 to 9.1, while with off-policy training, it rises from 16.3 to 18.6. Similarly, in Blackjack, non-Markovian prompts result in a higher average success rate.

In ALFWorld tasks, as demonstrated in~\cref{tab:markovian-results-alfworld}, the non-Markovian assumption yields marked gains in both average performance and out-of-distribution generalization. For instance, GFlowVLM with SubTB achieves an average success rate of 26.1 under the non-Markovian assumption compared to 22.1 under the Markovian setup. These results highlight the importance of historical context in improving task performance, particularly for challenging scenarios requiring long-term dependencies.

Interestingly, the non-Markovian assumption also benefits the baselines, including RL4VLM, resulting in a performance increase from 3.1 to 4.4 for Numberline for OOD tasks. This suggests that GFlowVLM is better equipped to leverage the additional context provided by non-Markovian prompts, enabling it to capture richer dependencies and improve both accuracy and diversity. Overall, the findings confirm that the non-Markovian assumption provides a more effective framework for reasoning-based tasks, particularly when combined with GFlowVLM's structured learning approach.

